\documentclass{article} % For LaTeX2e
\usepackage{iclr2027_conference,times}
\preprintcopy
\usepackage{amsmath,amsfonts,bm}

\def\eqref#1{equation~\ref{#1}}
\def\1{\bm{1}}

\DeclareMathAlphabet{\mathsfit}{\encodingdefault}{\sfdefault}{m}{sl}
\SetMathAlphabet{\mathsfit}{bold}{\encodingdefault}{\sfdefault}{bx}{n}

\usepackage{hyperref}
\usepackage{url}
\usepackage{graphicx}
\usepackage{amsmath}
\usepackage{booktabs}
\usepackage{subcaption}
\usepackage{wrapfig}
\usepackage{booktabs, arydshln, multirow}

\title{Beyond the Linear Representation Hypothesis: Non-Linear Activation Steering in Text-to-Image Models}

\author{Muhammad Atif Butt $^{1,2}$, Paweł Skierś $^{3,4}$, Joost Van De Weijer  $^{1,2}$, Kamil Deja $^{3,4}$ \\
$^{1}$ Computer Vision Center, Spain \\
$^{2}$ Computer Sciences Department, Universitat Autònoma de Barcelona, Spain \\
$^{3}$ Warsaw University of Technology, Poland \\
$^{4}$ IDEAS Research Institute, Poland \\
}

\begin{document}

\maketitle

\begin{abstract}
% \textcolor{red}{\textbf{[Atif --- initial version, to be revised]}} Mechanistic interpretability typically relies on Linear Representation Hypothesis (LRH), which assumes that high-level concepts are encoded as linear directions in activation space. However, this assumption remains underexplored in text-to-image diffusion models, particularly for continuous, ordered concepts where a single steering vector must remain valid across an entire semantic range. Concretely, we probe diffusion transformer activations under prompts placing an attribute at successive points along its range, taking the difference in means between adjacent points as each step's direction. Under LRH, these directions should be parallel and their path straight. However, consecutive steps align weakly, alignment from the start of the range stays well short of parallel, and the path departs substantially from the straight line joining its endpoints. To this end, we propose to model the traversal with Kolmogorov Arnold Network (KAN) layers, whose learned univariate functions allow the steering vector to vary with where along the range an attribute already resides. The resulting model fits the non-linear directions that a single linear direction fails to learn. We evaluate the proposed approach across several T2I diffusion transformers on a range of sequential and cyclic concepts, and find that nonlinear modeling fits the observed traversals more closely than a linear one.
Mechanistic interpretability often relies on the Linear Representation Hypothesis (LRH), which assumes that high-level concepts are encoded as linear directions in activation space. Yet a natural visual concept does not necessarily require a linear visual transition: between sunny and stormy lies an intermediate weather state such as a sky with a few white clouds, not simply a weaker storm; between a caterpillar and a butterfly, the progression is not a caterpillar with continuously growing wings. This raises the question of whether such true intermediate states are also represented nonlinearly by the model. Indeed, when we prompt text-to-image models directly for intermediate attributes, their activations rarely fall along the straight direction connecting the endpoints. Therefore, we propose KANSteer, which models concept traversal as a curve passing through its intermediate states. Seeking a representation that is both simple and interpretable, we propose to use Kolmogorov-Arnold Networks (KANs), which provide a one-dimensional coordinate whose learned functions define the trajectory. This allows the steering direction to vary along the concept while preserving an interpretable representation. Across several concepts and text-to-image diffusion transformers, we find that their activation trajectories substantially deviate from straight lines, and that KANSteer provide a closer fit and smoother traversal of intermediate attributes than linear steering.

\end{abstract}

\section{Introduction}

In deep learning, many semantic concepts admit an ordered progression, yet their representations need not change in a uniform, linear fashion. For example,
% in language models, the days of the week exhibit an ordered structure with a cyclical relationship between states rather than forming a simple infinite sequence~\citep{engels2024language}. Similar geometric complexity arises in the visual domain:
% , but with an important distinction as visual concepts are inherently continuous. 
the transition between seasons, the metamorphosis of an insect, or the progression of human aging is not simply a matter of scaling a single visual feature. Instead, these concepts evolve through meaningful intermediate states, where different attributes emerge, change, and disappear at different stages of the progression. Despite the complexity of these transitions, mechanistic interpretability has predominantly relied on the Linear Representation Hypothesis (LRH) to understand how models encode them. LRH posits that high-level concepts are represented by straight directions within a model's activation space. In text-to-image diffusion models, this perspective has motivated methods for precise control over generated content~\citep{surkov2024one}, model debiasing~\citep{modell2026probing} or unlearning~\citep{gaintseva2025casteer,cywinski2025saeuron}.
However, recent research in Large Language Models (LLMs) has highlighted a critical limitation of the LRH: while concepts may be linearly identifiable, they are not always linearly traversable. When navigating between related states--such as moving from Thursday to Friday or adjusting a character count-the underlying geometric path forms a curved manifold rather than a straight line~\citep{raval2026curveball,wurgaft2026manifold}.

This distinction becomes particularly important in visual domain. For a discrete structure such as days of the week, the goal is ultimately to reach one of a finite number of states; an imperfect trajectory can still place the representation close to the intended state. Visual concepts that are inherently continuous, offer no such simplification. Between two endpoints lie a sequence of meaningful intermediate states, and these states are themselves part of the concept. If they do not lie on a straight path, linear steering can therefore misrepresent the concept throughout the traversal, even when its endpoints are correctly identified. We directly test this by examining the activation trajectories induced by ordered visual attributes in text-to-image diffusion models. For each concept, we prompt the model with a sequence of progressively varying attribute levels and measure how their corresponding representations evolve in activation space. Across a range of visual concepts, these intermediate representations consistently deviate from straight line connecting the endpoints, revealing that visual concepts are not always linearly traversable.
% \kamil{Do we go back to the plot with values in Figure 1? This way we can put here some numbers?}

% This observation suggests that the difficulty lies less in representing a concept with a single degree of freedom than in requiring that degree of freedom to follow a fixed linear direction. 
This suggests that a concept can still have a simple 1D representation without following a fixed linear path for its intermediate states. We therefore model concept traversal as a one-dimensional curve through activation space. Using a Kolmogorov-Arnold Network (KAN)~\citep{liu2024kan0}, we parameterize this curve directly from the observed intermediate states, allowing steering direction to vary along trajectory while retaining a single interpretable coordinate. In this way, the representation preserves the simplicity of linear steering without imposing a straight path on a concept whose geometry may be inherently curved. We evaluate this approach across a range of ordered visual concepts and diffusion-model configurations. For each concept, we compare nonlinear steering against a linear trajectory under the same set of observed attribute states. While both approaches recover the endpoints, their behavior between them differs substantially: the nonlinear trajectory more closely follows the intermediate states in activation space and produces images that better match both the corresponding text and their prompt-generated visual references. These improvements are reflected consistently across directional and image-based similarity measures, indicating that the learned trajectory captures the progression between attributes rather than merely connecting its endpoints. We further test whether the learned trajectories can extend beyond the range of attributes used to fit them. By withholding later stages of each ordered concept during fitting, we evaluate whether the resulting trajectories can recover them. The nonlinear trajectories provide substantially more faithful extrapolations, both qualitatively and quantitatively, whereas linear steering remains constrained to a fixed direction and increasingly departs from the underlying progression.
%\kamil{From the samples I got it seems that it's degrading more, right? Let's confirm and rewrite it when we have final results}. 
% This demonstrates that modeling concept traversal as a nonlinear trajectory can capture structure that extends beyond the states directly observed during fitting.

Our contributions can be summarized as follows:
\begin{itemize}
\item We demonstrate that visual concepts in T2I diffusion models are not linearly traversable: intermediate attributes deviate from straight directions defined by their endpoints.
\item We introduce KANSteer, a 1D-nonlinear steering framework that replaces the fixed steering direction with a learned activation-space trajectory, preserving an interpretable scalar concept coordinate while allowing the steering direction to vary along the trajectory.
\item Given the same supervision, KANSteer improves over the linear direction on every metric and all three model configurations, raising directional CLIP from $0.108$ to $0.170$ and lowering FID from $47.9$ to $28.6$ on FLUX.1-schnell (1-step). Notably, it transfers unchanged from the FLUX.1-schnell 1-step activations it is fitted on to FLUX.1-schnell 4-step and FLUX.1-dev generation. Furthermore, we show that KANSteer can extrapolate beyond its fitted range while generating realistic samples.
% \item Given the same supervision, KANSteer improves over the linear direction on every metric and all three model configurations, raising directional CLIP from $0.108$ to $0.170$ and lowering FID from $47.9$ to $28.6$ on FLUX.1-schnell (1-step). Notably, it transfers unchanged from the FLUX.1-schnell 1-step activations it is fitted on to FLUX.1-schnell 4-step and FLUX.1-dev generation. Furthermore, we show that KANSteer can extrapolate beyond its fitted range while generating realistic samples.
% \item Finally, the learned trajectory extrapolates beyond the range it is fitted on, where KANSteer achieves better directional CLIP ($0.050 \rightarrow 0.110$) relative to the linear direction, which instead overshoots off the natural image distribution. The displacements it predicts agree with the model's own at cosine $0.55$ versus $0.37$ for the best line, and the margin is largest on the concepts whose trajectories bend the most.

% \item We show that nonlinear steering more faithfully recovers intermediate attributes and extrapolates to unseen stages of a concept.
\end{itemize}

\section{Related Work}

\paragraph{Linear representation hypothesis (LRH).} 
The LRH proposes that a high-level concept is represented by a direction in activation space. This idea has been formalized for language models by~\citet{park2023linear,park2024geometry,jiang2024origins}, and used to control properties such as truthfulness~\citep{li2023inference0time}, sentiment~\citep{tigges2023linear}, and refusal~\citep{arditi2024refusal} by adding a fixed steering vector. An analogous picture has emerged in vision, where diffusion models exhibit surprisingly linear internal representations. Simple linear probes can extract scene properties--such as depth and figure-ground structure--from intermediate activations well before denoising is complete~\citep{chen2023beyond}. Similarly, linear projections of attention layers in diffusion transformers yield sharp concept saliency maps~\citep{helbling2025conceptattention0}. Recent work highlights that specific knowledge localizes to just a few blocks~\citep{basu2024mechanistic,basu2024-localizing-knowledge,NEURIPS2025_157a6107}, and sparse autoencoders can decompose these activations into interpretable, causal features~\citep{surkov2024one,cywinski2025saeuron}.

These structural discoveries naturally motivated direct interventions. Early methods identified the U-Net bottleneck as a semantic latent space that is largely homogeneous and linear, where scaling a single vector gradually alters image attributes~\citep{DBLP:conf/iclr/KwonJU23}. Subsequent approaches extracted these directions unsupervised~\citep{dalva2023noiseclr0} or used them to guide attribute distributions~\citep{DBLP:conf/cvpr/PariharBBMKB24}. A second line of research directly ported activation steering from language models to image generators. In these methods, steering directions are computed via optimal transport~\citep{rodriguez2025controlling,rodriguez2025lineas} or by isolating the mean difference between prompts with and without a specific concept~\citep{gaintseva2025casteer}. Other approaches apply this linear logic elsewhere, calculating directions in the text embedding space~\citep{kim2025concept,baumann2024continuous0} or applying continuous attribute control via learned low-rank updates~\citep{DBLP:conf/eccv/GandikotaMZTB24, Gandikota_2025_ICCV}. Yet, despite the variety of these activation-space methods, underlying intervention mechanism remains identical: they all steer by updating $x' = x + \alpha v$ for a fixed vector $v$.

\paragraph{Extending linear representation hypothesis}
In LLMs, several works argue that the linear manifold does not hold for all concepts. The clearest example is the representation of weekdays, where each day is identifiable as a separate direction but the seven form a circle rather than a line, so moving from Monday to Tuesday requires a different displacement than moving from Thursday to Friday~\citep{engels2024language}. Character counts likewise lie on curved low-dimensional manifolds~\citep{gurnee2026models}. A concept can therefore be \emph{linearly identifiable} without being \emph{linearly traversable}.
This has motivated steering methods that replace a fixed direction with a trajectory, learning a concept manifold with a supervised probe~\citep{modell2026probing}, obtaining curved trajectories with polynomial kernel PCA~\citep{raval2026curveball}, or fitting cubic splines through concept centroids, which keeps steering on trajectories already present in the model where a linear direction passes through off-manifold regions and produces unnatural outputs~\citep{wurgaft2026manifold}. \citet{engels2024language} note that such representations are ''mostly discontinuous'', as the model does not naturally represent states between adjacent weekdays or months. Visual attributes differ, since meaningful intermediates are expected even when only a few are specified, and in T2I diffusion models they can be generated and observed directly. Unlike prior approaches that recover trajectories from activation geometry or concept centroids, we fit a parametric trajectory with a KAN to the intermediate states themselves, which gives an explicit scalar coordinate and a channel-wise decomposition, and makes the trajectory define the steering rule rather than being fitted separately from the intervention.

\begin{figure}[t]
    \begin{center}
        \includegraphics[width=\linewidth]{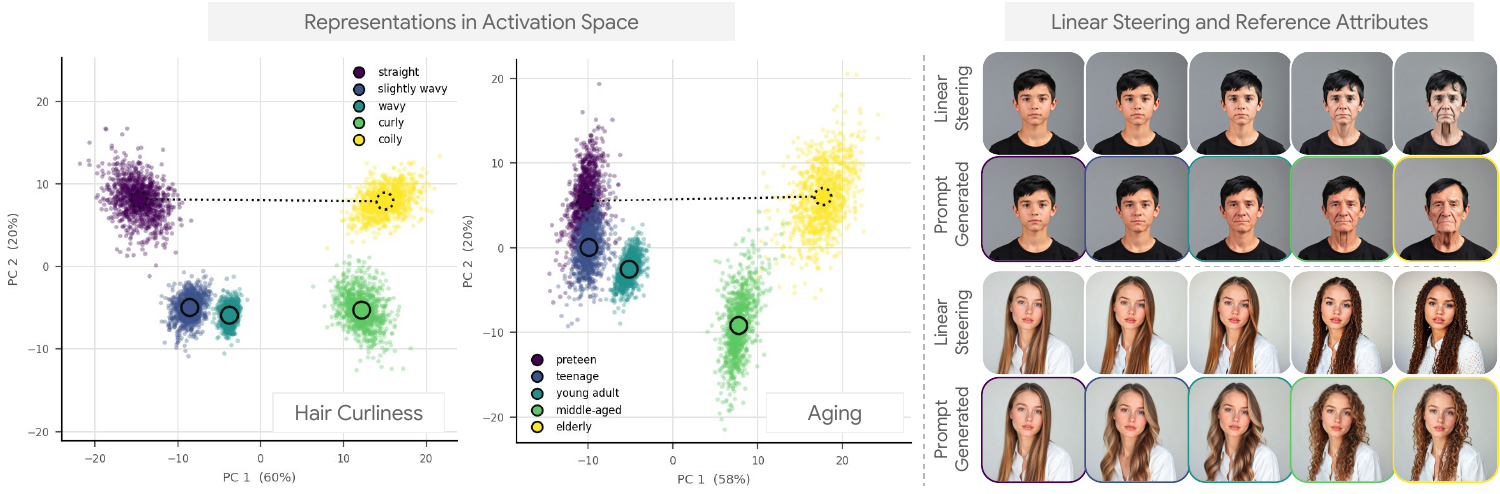}
    \end{center}
    \caption{Many concepts do not vary linearly in activation space. \textbf{Left:} PCA projection of block-$q$ delta activations for three concepts (see more in Appendix~\ref{fig:pca_app}); each point is one seed, with attribute centroids, and the two endpoints outlined. The intermediate attributes bend away from straight line between the endpoints, so a fixed direction cannot pass through these attributes. \textbf{Right:} Sweeping closed-form linear direction (\emph{Linear Steering}) holds the source attribute and then jumps to the target, while prompt-generated attributes it is fitted on (\emph{Prompt Generated}) traverse the concept smoothly. }
    %\kamil{Can we also add points on the linear trajectory? and aling those legends to bottom left e.g.}}
    %\kamil{Can we add here a line between endpoints and 5 points on this line following the colors we already have. For the generations from prompts maybe let's put them in small frames from following the colors of their embeddings? Pictures are also much too small, most of the reviewers will read it on their laptops, and when I open the paper on the laptop screen those individual images are 1cm x 1cm and you cannot see any differences in faces or hair. I'm not sure how to make it bigger, maybe only 2 concepts + separately plot that we had before for multiple concepts?} \pawel{+1 for all of that}}
    \label{fig:motivation}
\end{figure}

% \kamil{Mention that Engels says that they representations are "mostly discontinuous" so for example nothing lies between adjacent weekdays by default. This is contrary to vision where we expect continuity and we only define some intermediate points}

% \paragraph{Activation steering in DiTs.}
% CASteer, SAE-based steering, difference-in-means steering, Concept Sliders, ...

% \paragraph{Interpretability in diffusion transformers.}

\section{Concept Trajectories are not Linear in Activation Space}
\label{sec:motivation}
Before moving to the proposed method, we first ask a simple question: if we learn a linear steering direction between two endpoints, can this direction reproduce the intermediate states of the concept? A common approach to activation steering in T2I models is to define a concept using two attributes and learn a direction between them~\citep{gaintseva2025casteer,rodriguez2025controlling}. Under the linear representation hypothesis (LRH), moving along this direction should then traverse the concept from one endpoint to the other, passing through the corresponding intermediate states.

% \subsection{Setup}

We test this directly by constructing intermediate states independent of steering direction. For each concept, we specify five ordered attributes $a_1 < a_2 < \cdots < a_K$, with first and last serving as the steering endpoints and the remaining three defining intermediate targets. For example, for hair curliness we use \textit{straight}, \textit{slightly wavy}, \textit{wavy}, \textit{curly}, and \textit{coily}. For aging, we use \textit{preteen}, \textit{teenage}, \textit{young adult}, \textit{middle-aged}, and \textit{elderly}. We then generate images for all five attributes using same prompt template, and same seed, changing only attribute phrase (See Section~\ref{sec:experimental_details} for more details). These prompt-generated representations give us an explicit trajectory of intermediate states against which to evaluate linear steering. For each concept, we first learn the linear steering direction using only the two endpoint attributes. We then sweep the steering coefficient between these endpoints and compare the resulting representations to those obtained from the three intermediate prompts. % as presented in Fig.~\ref{fig:motivation}. 
To assess whether the ordered attributes of a concept lie on a straight path, here we analyze the geometry of the attributes in the activation space. For each concept we take the delta activation of every \{attribute, seed\} pair, and since all $K$ attributes of a seed are generated from the same initial noise, we subtract the per-seed mean across attributes. This removes the shared initial noise and leaves the variation which the attributes induce. We then project the centered activations onto their leading principal components. If the attributes are linearly encoded, the $K$ attribute clusters would fall on a single line in this projection, and their ordering would run straight along it. However, the results in Figure~\ref{fig:motivation} reveal that across the concepts, the intermediate attributes sit off the straight path between the two endpoint attributes. A quantitative analysis confirms that intermediate attributes deviate substantially from the straight path between the endpoints (Appendix~\ref{app_sec:motivation}).

\begin{figure}[t]
    \begin{center}
        \includegraphics[width=\linewidth]{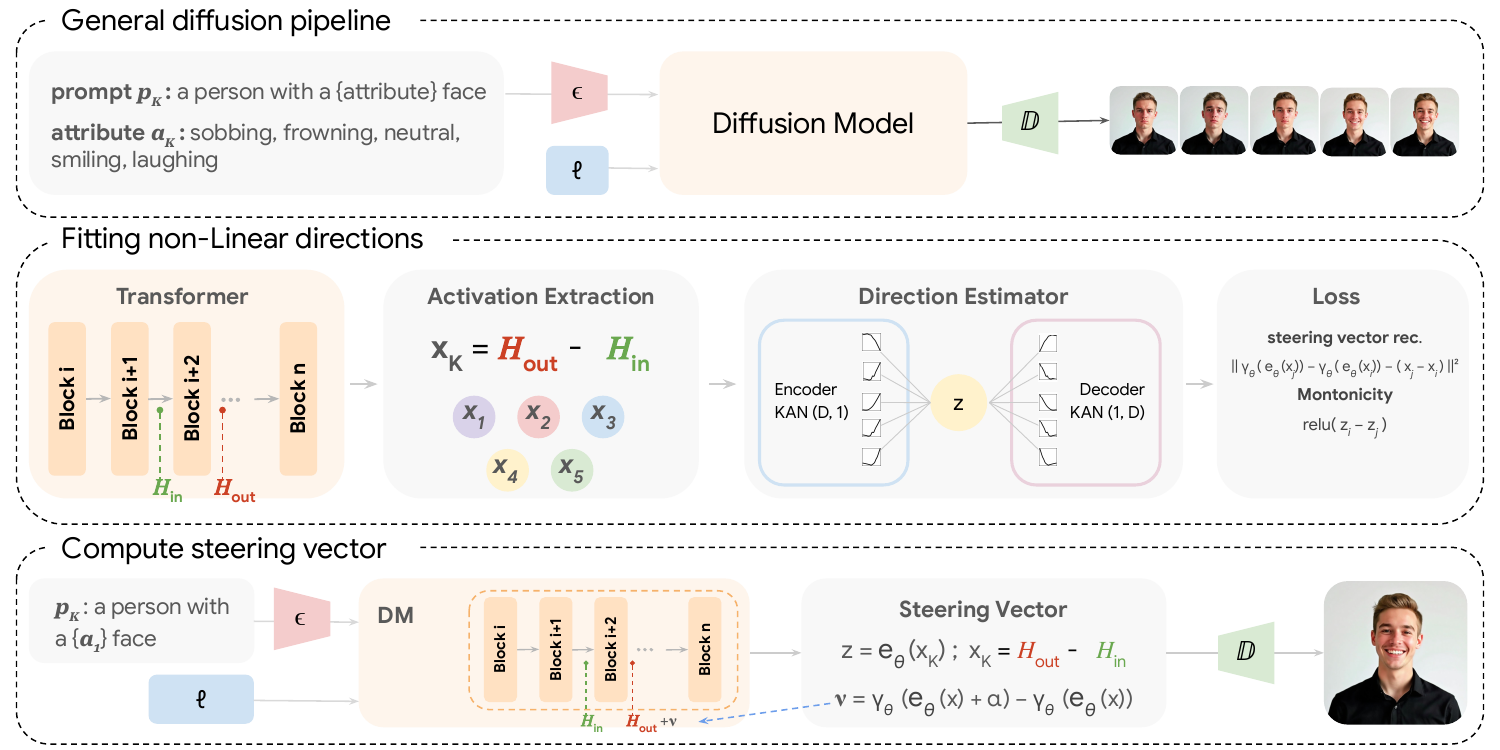}
    \end{center}
    \caption{Overview of KANsteer. \textbf{Top:} Each concept is instantiated at $K=5$ ordered attributes ($a_1,..,a_k$), and generated all from same initial noise. \textbf{Middle:} At block $i$, we compute residual activation from the image stream. Every attribute serves as a source $x_i$ and later one as its target $x_j$, giving source–target combinations, each a training example with source $X = x_i$ and difference $Y_{ij} = x_j - x_i$. A 1D autoencoder maps $x_i$ to a coordinate $z_k = e_\theta(x_i)$ and recovers the direction as a difference of decoded positions, fitted with $\mathcal{L}_{\mathrm{rec}}$ and $\lambda
\sum_{i<j}
\max(0,z_i-z_j)$ to ensure reconstruction and monotonicity of the attributes. \textbf{Bottom:} At inference, the coordinate is and shifted by $\alpha$, so the steering vector computed $\nu_\alpha(x)
=
\gamma_\theta\left(e_\theta(x)+\alpha\right)
-
\gamma_\theta\left(e_\theta(x)\right)$.}
    \label{fig:method}
\end{figure}

\section{Method}
\label{sec:method}

In Section~\ref{sec:motivation}, we establish that the attributes of many concepts do not lie on a line in activation space, so a single direction cannot carry a concept through its intermediate states. This motivates replacing fixed steering direction used by linear activation steering with a 1D trajectory. To do so, we fit a model of how activations change across all $K$ attributes of a concept and derive the steering rule from that model. Steering then amounts to moving the activation along this learned trajectory.

\subsection{A one-dimensional trajectory model}
\label{sec:method_trajectory}

We model the activations induced by the ordered attributes as points along a shared \emph{concept trajectory} in activation space. For each seed, we generate one image for each attribute $a_1,\ldots,a_K$ from the same initial noise, changing only the attribute phrase. Because the resulting images share their initial noise, their activations retain much of the same image-specific content, such as layout, identity, and background. We therefore view the activation under attribute $a_k$ as
\begin{equation}
x_k \approx b + \gamma(t_k),
\label{eq:trajectory_model}
\end{equation}
where $b\in\mathbb{R}^D$ is the content shared across the attributes of a given seed and patch, and
$\gamma:\mathbb{R}\rightarrow\mathbb{R}^D$ is the \emph{concept trajectory}. The trajectory describes how the representation changes as the attribute progresses, while the scalar $t_k$ specifies the position of attribute $a_k$ along that trajectory.

A displacement from $x_i$ to $x_j$ can be expressed as $\gamma(t_j)-\gamma(t_i)$. Thus, steering requires two components: a model of the trajectory $\gamma$, and a way to assign an activation its coordinate on that trajectory.

\subsection{Fitting the trajectory}
\label{sec:method_fitting}

Given this formulation, we jointly learn trajectory and the coordinate assigned to each activation. Let
$e_\theta:\mathbb{R}^D\rightarrow\mathbb{R}$ map an activation to its coordinate on the trajectory, and let
$\gamma_\theta:\mathbb{R}\rightarrow\mathbb{R}^D$ map a coordinate to the corresponding point in activation space. For every seed and patch, each ordered pair of attributes provides a transition that the trajectory must explain. For a pair $(i,j)$ with $i<j$, the observed transition is $Y_{ij}=x_j-x_i$. The model predicts the corresponding displacement as,
\begin{equation}
\gamma_\theta\left(e_\theta(x_j)\right)
-
\gamma_\theta\left(e_\theta(x_i)\right).
\end{equation}
We therefore minimize,
\begin{equation}
\mathcal{L}_{\mathrm{rec}}
=
\sum_{i<j}
\left|
\gamma_\theta\left(e_\theta(x_j)\right)
-
\gamma_\theta\left(e_\theta(x_i)\right)
-
(x_j-x_i)
\right|_2^2,
\label{eq:rec}
\end{equation}
over all source--target pairs. Because the reconstruction loss depends only on differences along the trajectory, it does not by itself enforce that the learned coordinates follow the semantic ordering of the attributes. We therefore add a monotonicity penalty for the violations of the desired ordering.
\begin{equation}
\mathcal{L}
=
\mathcal{L}_{\mathrm{rec}}
+
\lambda
\sum_{i<j}
\max(0,z_i-z_j),
\qquad
z_k=e_\theta(x_k),
\label{eq:monotone_loss}
\end{equation}

\subsection{Steering along the learned trajectory}
\label{sec:method_steering}

Once the encoder and trajectory have been learned, steering requires no separate vector estimation. Given an activation $x$, its current position on the trajectory is $z=e_\theta(x)$. A steering strength $\alpha$ specifies a displacement in this one-dimensional coordinate. We therefore move from $z$ to $z+\alpha$ and apply the corresponding displacement in activation space:
\begin{equation}
\nu_\alpha(x)
=
\gamma_\theta\left(e_\theta(x)+\alpha\right)
-
\gamma_\theta\left(e_\theta(x)\right),
\qquad
x' = x + \nu_\alpha(x).
\label{eq:steering}
\end{equation}

The steering vector $\nu_\alpha(x)$ depends on current position $e_\theta(x)$. Thus, same coordinate shift can produce different activation-space displacements at different points along the concept. This allows the intervention to follow a curved trajectory while retaining a single scalar notion of concept strength.

Linear steering is recovered as a special case. If the trajectory is affine, $\gamma_\theta(z)=c+zw$, then
\begin{equation}
\nu_\alpha(x)
=
[c+(z+\alpha)w]-[c+zw]
=
\alpha w,
\end{equation}
which is independent of $x$. Thus, the proposed formulation generalizes fixed-vector steering: a straight trajectory recovers the standard linear intervention, while a curved trajectory produces position-dependent steering vectors.

\subsection{KAN parameterization}
\label{sec:method_kan}

The trajectory and encoder can be parameterized by many nonlinear one-dimensional models. We use Kolmogorov--Arnold Networks (KANs)~\citep{liu2024kan0} because they preserve an explicit scalar coordinate while allowing the trajectory to be nonlinear.

The encoder $e_\theta:\mathbb{R}^{D}\rightarrow\mathbb{R}$ is implemented as a one-dimensional KAN that maps the activation to the scalar coordinate $z$. The decoder $\gamma_\theta:\mathbb{R}\rightarrow\mathbb{R}^{D}$ is implemented as a KAN layer of width $[1,D]$. In this form, each output coordinate is a learnable univariate function of the same scalar input:
\begin{equation}
\gamma_\theta(z)
=
\left[
\phi_1(z),\ldots,\phi_D(z)
\right],
\end{equation}
with each function parameterized as
\begin{equation}
\phi_d(z)
=
w_{b,d},\mathrm{SiLU}(z)
+
w_{s,d},\mathrm{Spline}_d(z).
\label{eq:kan}
\end{equation}
The spline term provides flexibility, while SiLU provides a smooth base function. This allows to deviate from linearity and represent curved activation-space trajectories observed in Section~\ref{sec:motivation}.

\section{Experiments}

\subsection{Experiment Setup}
\label{sec:experimental_details}

\paragraph{Models.} We evaluate KANSteer on three distinct FLUX.1 configurations. FLUX.1 employs a transformer-based architecture comprising 19 dual-stream and 38 single-stream blocks, conditioned on T5-XXL and CLIP text encoders. We contrast FLUX.1-schnell at both 1 and 4 sampling steps against FLUX.1-dev at its native high-step regime. This design isolates the variable of denoising-step budget while varying the distillation objective between the schnell and dev variants.

\paragraph{Delta activations at a residual block.}
% \kamil{Moved from another section, align it with this one}
% \kamil{This might be important but somewhere in experimental setup?} \pawel{important but not in the "concepts are not linear" subspace. Move to method/preliminaires/exp setup.}
We analyze a text-to-image diffusion transformer where the backbone is a stack of residual blocks operating on a stream of $P$ image patches of width $D$. Let $H^{\ell} \in \mathbb{R}^{P \times D}$ denote the image stream entering block $\ell$, and $\mathcal{B}_{\ell}$ the transformation it applies under text conditioning $c$. Each block writes an additive update to the stream,

\begin{equation}
    H^{\ell+1} \;=\; H^{\ell} \;+\; \mathcal{B}_{\ell}\!\left(H^{\ell}, c\right),
    \label{eq:block}
\end{equation}
and our unit of analysis is that update at a single patch $p$,
\begin{equation}
    x_p^{\ell} \;:=\; \mathcal{B}_{\ell}\!\left(H^{\ell}, c\right)_{p}
    \;=\; H^{\ell+1}_{p} - H^{\ell}_{p} \;\in\; \mathbb{R}^{D},
    \label{eq:delta}
\end{equation}

which isolates what the block contributes from what the residual stream already carried. We refer to $x_p^{\ell}$ as a \emph{delta activation} and drop the block index hereafter. All analysis is performed at a single block $q$, selected as the most consequential for the attributes we study; the selection procedure, based on activation norms and activation patching, is detailed in Appendix~\ref{app_sec:localize_block}.

\begin{figure}[t]
    \begin{center}
        \includegraphics[width=\linewidth]{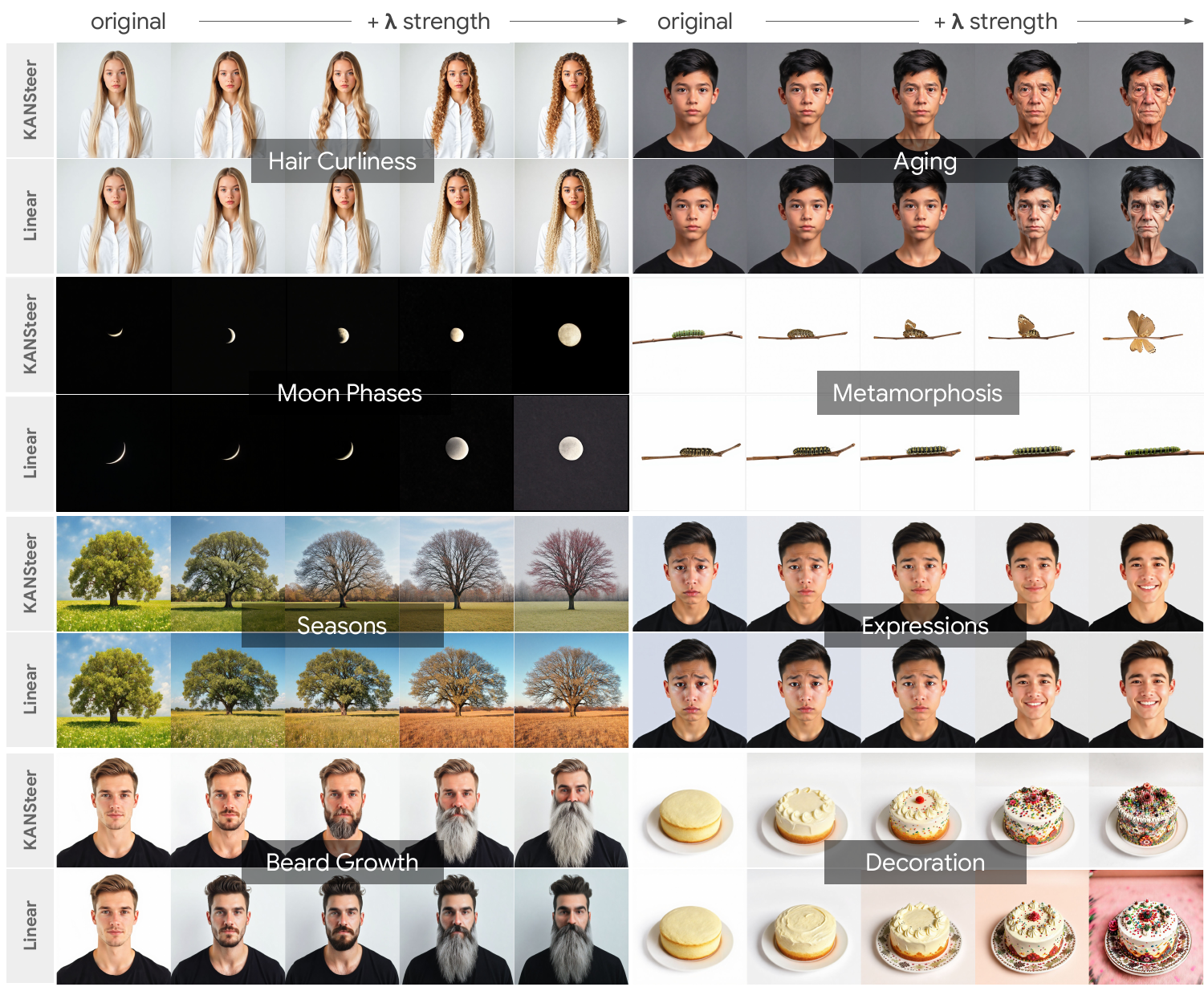}
    \end{center}
    \caption{Qualitative comparison on FLUX.1-schnell. We sweep the steering strength from unsteered generation, comparing the fixed direction against KANsteer learned trajectory. \textit{KANSteer traverses the attribute gradually, with intermediate columns showing intermediate attributes, while the linear direction holds the source attribute and then transitions abruptly.}}
    \vspace{-3mm}
    \label{fig:qual_results}
\end{figure}

\paragraph{Implementation Details.} We generate at $1024^2$, giving $P = 4096$ image patches of width $D = 3072$. All fitting is done at block 18, the last dual-stream block, which we identify as the most influential for these attributes by activation norm and causal patching (Appendix~\ref{app_sec:localize_block}), where we extract the image-stream delta activation $x = H_{out} - H_{in} \in \mathbb{R}^{P \times D}$. Following~\citep{surkov2024one}, we fit on a random $10\%$ of the patches per seed, using the same patch indices across the $K$ attributes of a seed so that the pairing in \eqref{eq:rec} is preserved, with $800$ training and $100$ validation seeds. Since KAN layers place their spline grid over $[-1,1]$, we divide every activation by the mean patch norm before fitting, and set the grid to $5$ intervals with order-$3$ splines. A few patches carry massive norm but no concept information and would dominate a squared-error fit, so we drop any patch whose norm exceeds the $0.99$ quantile of the per-seed patch norms, taken as the maximum over the $K$ attributes (Appendix~\ref{app_sec:localize_block}). At inference we add the steering vector at every block from 18 onward~\citep{surkov2024one}, rescaled by the mean patch norm to each block's activation scale. 

\paragraph{Evaluation Metrics.} We evaluate on a held-out test set of $600$ seeds per attribute, $3000$ per concept. All metrics are reported at a per-attribute strength $\alpha_k$ chosen on a calibration split, as the value that maximizes SigLIP2~\citep{tschannen2025siglip} similarity between steered image and attribute prompt, subject to the attribute order $a_1 < \dots < a_K$. \textbf{Directional-CLIP}~\citep{gal2022stylegan}, $\cos\!\big(\phi(I_{\alpha_k}) - \phi(I_0),\, t_k - t_0\big)$, asks whether steering moves the image in the intended direction; \textbf{CLIP-Image}, $\cos\!\big(\phi(I_{\alpha_k}), \phi(\mathrm{nat}_k)\big)$, is its alignment with direct prompting for the same attribute and seed; and \textbf{CLIP-T2I}, $\cos\!\big(\phi(I_{\alpha_k}), t_k\big)$, its agreement with attribute prompt. \textbf{LPIPS}~\citep{zhang2018unreasonable} is the perceptual distance to $I_0$, capturing change beyond the target attribute, and \textbf{FID}~\citep{heusel2017gans}, pooled over attributes against the prompt-generated distribution, reports if steering stays on the natural manifold. \textbf{FVU} reports how much of the model's own attribute displacement the fitted curve fails to reconstruct on held-out activations.

\subsection{Results} 
We demonstrate results in figure~\ref{fig:qual_results}. The fixed direction holds source attribute the most and then switches abruptly to the target. For instance, hair stays straight and then curls, and a young face becomes elderly. KANSteer advances through the attribute continuously and recovers intermediate states, giving wavy hair before curly, and a face that passes through middle age. The same pattern holds across the concepts (Appendix~\ref{app_sec:qual_res}). Where the attributes lie close to a line i.e., human age, the linear direction still reaches the target and the both methods differ mainly in continuous traversal. Where the trajectory bends, as for facial expression, hair curliness and butterfly metamorphosis, it skips intermediate attributes and forces the endpoint at the cost of leaving reference distribution.

Table~\ref{tab:linear_kan_comparison} reports the metrics averaged over the concepts, with per-concept results in Appendix~\ref{app_sec:quan_res}. KANSteer improves on every metric at every setting. It follows the target direction more closely (directional CLIP $0.108 \rightarrow 0.170$ at 1-step and $0.053 \rightarrow 0.110$ on dev) and stays nearer the prompt-generated distribution (FID $47.9 \rightarrow 28.6$ and $46.7 \rightarrow 33.8$). The FID gap is the quantitative counterpart of the sweeps, where the linear direction reaches the target attribute only by driving the image off the natural distribution. The advantage is not specific to single-step sampling, as directional CLIP rises from $0.086$ to $0.149$ and FID falls from $38.2$ to $28.3$ at 4 steps. KANSteer also reaches these attributes with less collateral change to the image, lowering LPIPS from $0.243$ to $0.228$ at 1-step and from $0.212$ to $0.177$ on dev, so the gain does not come from perturbing the activation more strongly. We fit the trajectory once on 1-step activations and reuse it unchanged at 4 steps and on FLUX.1-dev, which takes 50 steps, and in activation space the curve leaves less of the model's own displacement unexplained than the best line at every setting, $0.467$ against $0.619$ at 1-step and $0.481$ against $0.544$ on dev. Per concept, the two methods coincide at the source attribute and separate as steering progresses. Human age on FLUX.1-dev is the clearest case, where at the middle attribute KANSteer lowers LPIPS from $0.493$ to $0.142$ and raises directional CLIP from $0.137$ to $0.241$. On facial expression the linear direction is even slightly negative at the neutral attribute ($-0.010$ against $0.093$), moving the image away from the state it should pass through. The concepts that gain most are those whose trajectories bend, matching the qualitative split.

\begin{table}[t]
\centering
\caption{Comparison of Linear steering and KANSteer on FLUX.1-schnell (1- and 4-step) and FLUX.1-dev, under two fitting regimes. \emph{In-range}: the curve is fitted and steered on all ordered attributes. \emph{Extrapolation}: the curve is fit on the first three attributes and steered over all five; Directional CLIP, CLIP-Image, and CLIP-T2I measure alignment with the target attribute; LPIPS and FID measure fidelity to the unsteered image and to the prompt generated distribution; FVU measures how well the fitted curve reconstructs the model's own attribute displacement in activation space.}
\resizebox{\linewidth}{!}{%
\begin{tabular}{llcccccc}
\toprule
\textbf{Model} & \textbf{Method} &
\textbf{dir-CLIP $\uparrow$} & \textbf{CLIP-I $\uparrow$} & \textbf{LPIPS $\downarrow$} & \textbf{CLIP-T2I $\uparrow$} & \textbf{FID $\downarrow$} & \textbf{FVU $\downarrow$} \\
\midrule
\multicolumn{8}{c}{\textit{In-range (fit: 5 attributes, steer: 5)}} \\
\midrule

%Linear	0.1079425	0.856325	0.24345	0.5367175	47.86	0.6187
%KAN	0.17035	0.901775	0.227525	0.5592725	28.56	0.4674

\multirow{2}{*}{FLUX.1 Schnell (1-step)} & Linear   & 0.1079 & 0.8563 & 0.2434 & 0.5367 & 47.86 & 0.6187 \\
                                         & KANSteer & 0.1703 & 0.9017 & 0.2275 & 0.5592 & 28.56 & 0.4674 \\

% linear	0.08553125	0.8147625	0.22439375	0.505654375	38.176	0.4906
% kan	0.14931875	0.8625875	0.193275	0.53370125	28.256	0.4531
\cdashline{1-8}
\multirow{2}{*}{FLUX.1 Schnell (4-step)} & Linear   & 0.0855 & 0.8147 & 0.2243 & 0.5056 & 38.17 & 0.4906 \\
                                         & KANSteer & 0.1493 & 0.8625 & 0.1932 & 0.5337 & 28.25 & 0.4531 \\
\cdashline{1-8}

% linear	0.0534	0.793265	0.21211	0.5006655	46.733	0.5437
% kan	0.11016	0.8375075	0.176655	0.5320815	33.793	0.4811

\multirow{2}{*}{FLUX.1 dev}              & Linear   & 0.0534 & 0.7932 & 0.2121 & 0.5006 & 46.73 & 0.5437 \\
                                         & KANSteer & 0.1102 & 0.8375 & 0.1766 & 0.5320 & 33.79 & 0.4811 \\

% Linear	0.0501	0.6114	0.3754	0.4469	89.12	0.6938
% KAN	0.1104	0.741	0.2971	0.482	61.74	0.547

\midrule
\multicolumn{8}{c}{\textit{Extrapolation (fit: attributes 1--3, steer: 1--5)}} \\
\midrule
\multirow{2}{*}{FLUX.1 Schnell} & Linear   & 0.0501 & 0.6114 & 0.3754 & 0.4469 & 89.12 & 0.6938 \\
                                         & KANSteer & 0.1104 & 0.741 & 0.2971 & 0.482 & 61.74 & 0.547 \\
% \cdashline{1-7}
% \multirow{2}{*}{FLUX.1 Schnell (4-step)} & Linear   & x & x & x & x & x & \\
%                                          & KANSteer & x & x & x & x & x & \\
% \cdashline{1-7}
% \multirow{2}{*}{FLUX.1 dev}              & Linear   & x & x & x & x & x & \\
%                                          & KANSteer & x & x & x & x & x & \\
\bottomrule
\end{tabular}%
}
\label{tab:linear_kan_comparison}
\end{table}

\subsection{Steering beyond the fitted range}

\setlength{\columnsep}{4pt}
\begin{wrapfigure}{r}{0.55\linewidth}
\vspace{-14pt}
\centering
\includegraphics[width=\linewidth]{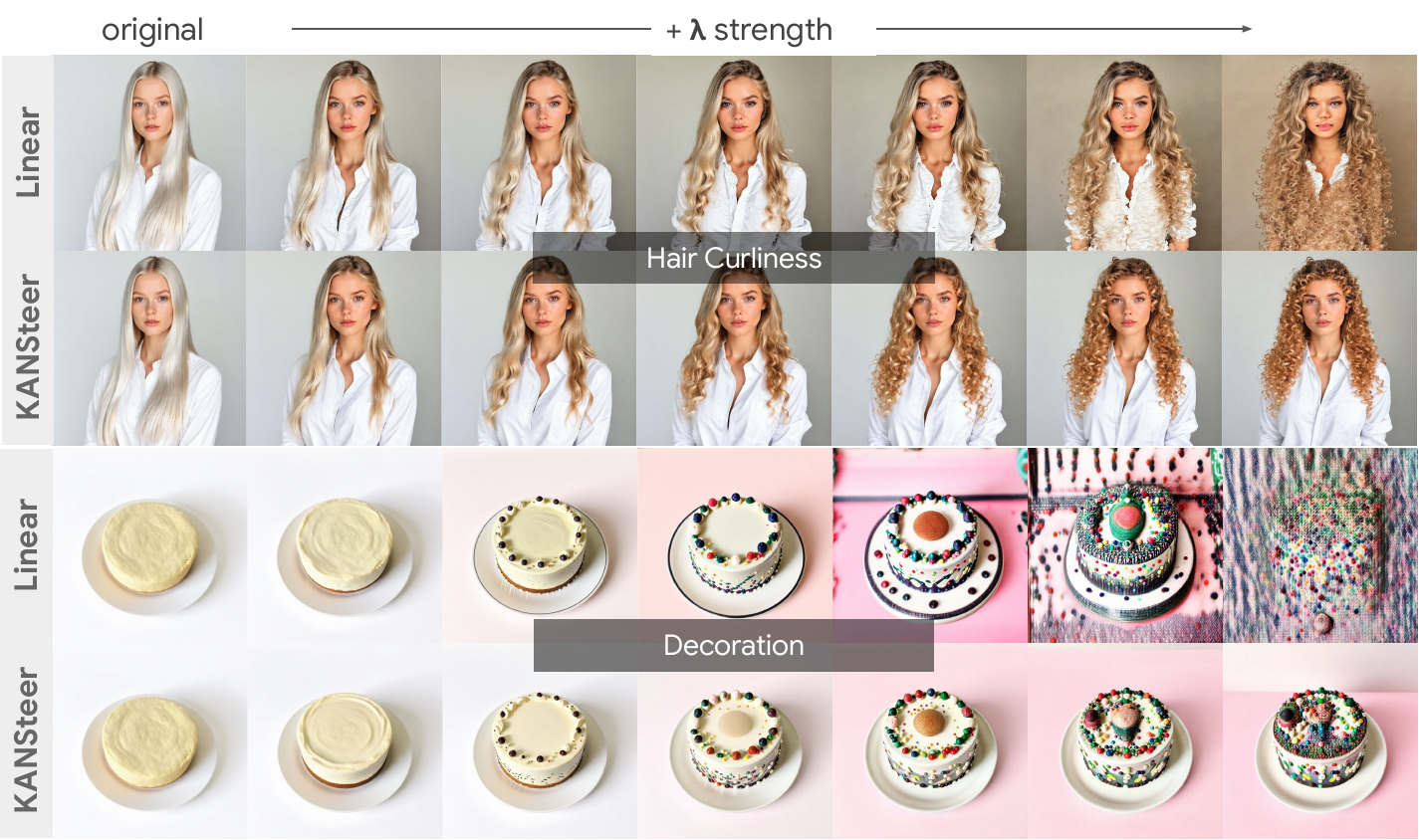}
\vspace{-12pt}
\caption{Extrapolating linear and KANSteer directions, both are fitted on first three attributes ($a_1, a_2, a_3$) and steered over all five attributes.}
%KANSteer continues the trajectory smoothly into the unseen range, while the linear direction overshoots into off-manifold artifacts.
\label{fig:steering_extrapolation}
\end{wrapfigure}

The in-range results fit the trajectory on all five attributes. A natural follow-up is whether a trajectory fitted on a partial range of attributes can extrapolate to unseen ones. For instance, if a curve is fitted on first three attributes of a concept, can steering recover the remaining which are not observed during training? We train both models on $a_1, a_2, a_3$ and evaluate over $a_1, \dots, a_5$, using same held-out seeds and metrics as the in-range setting. For the linear model, extrapolation simply extends the fitted line; the direction is unchanged, so the same hold-and-jump failure that Section~\ref{sec:motivation} diagnosed persists at the same steering strengths. For the KANSteer, the B-spline grid covers only the coordinate range seen during training; beyond it the decoder reduces to its SiLU base, $w_b\,\mathrm{silu}(z)$, which provides a smooth monotone continuation rather than a flat cutoff. Figure~\ref{fig:steering_extrapolation} shows that on hair curliness, KANSteer trained on straight through wavy still produces curly and coily hair at higher steering strengths, while the linear direction overshoots into artifacts. On decoration, same pattern holds where KANSteer advances gradually from plain to elaborately decorated, whereas the linear direction drives the image off the natural distribution. The quantitative comparison in Table~\ref{tab:linear_kan_comparison} also confirms that KANSteer retains its advantage on unseen attributes, indicating that the fitted curve captures a continuous structure in activation space rather than memorizing training points.

\subsection{Does nonlinear steering better predict the concept trajectory?}

\begin{figure}[t]
    \begin{center}
        \includegraphics[width=\linewidth]{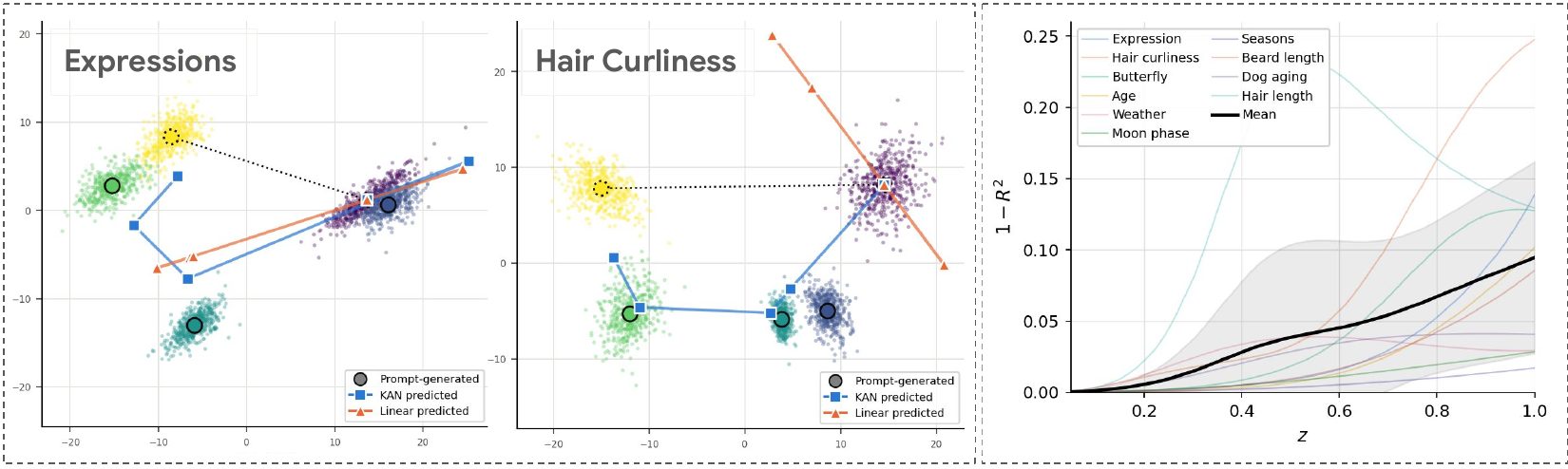}
    \end{center}
    \caption{\textbf{Left:} PCA of block-18 activation, with per-seed scatter (dots), and model predicted centroids  overlaid. \textit{KANSteer predictions follow the curved centroid path whereas the linear predictions collapse onto a single direction that misses the curvature.} \textbf{Right:} Non-linearity $(1 - R^2)$ of KANSteer decoder curve as a function of the z interval, measured by the fraction of decoder variance not captured by the best linear fit. \textit{Over the full attribute range the non-linear share rises to $10-25\%$ on concepts such as hair curliness and facial expression.}}
    \label{fig:better_steering}
    \vspace{-3mm}
\end{figure}

The previous sections show that KANSteer outperforms linear steering on reconstruction and perceptual quality. Here we ask a further question that does the nonlinearity in the KAN decoder actually capture the geometry of the concept trajectory? To realize this, we first measure the non-linearity of the learned deocder curve $\gamma(z)$. The decoder function of $\phi_c(z)$ is evaluated over the empirical z-range and decomposed against orthonormal polynomials; the nonlinear share $1 - R^2$ is the fraction of centered energy above degree one. Figure~\ref{fig:better_steering}(Right) demonstrates $1 - R^2$ as a function of sub-interval width which shows that over the full attribute range, the nonlinear share grows monotonically, reaching $25\%$ on hair curliness and $14\%$ on facial expression. This is consistent with a trajectory that bends gradually over the course. Secondly, we test whether this curvature predicts where concepts actually land in activation space. For each held-out seed we encode all five attribute activations through the trained model, compute the predicted displacement $\gamma_\theta\left(e_\theta(x_j)\right)
- \gamma_\theta\left(e_\theta(x_i)\right)$, and average over seeds and patches to obtain centroid-level predictions. We then measure the cosine similarity between each model's predicted centroid displacement and the actual displacement. Across the concepts, KANSteer achieves a mean cosine alignment of 0.55 against 0.37 for the linear model. The gap is largest on hair curliness (0.88 vs. 0.20), where the trajectory curves the most. Figure~\ref{fig:better_steering}(Left) shows per-seed patch-mean activations projected onto first two principal components. For each held-out seed, KANSteer and the linear model predict per-patch displacements across all five attributes; averaging these predictions over seeds and patches yields predicted centroids which shows that KANSteer follow the curved path traced by the actual prompt-generated distribution (circles), while the linear-direction collapse onto a single direction that diverges from the true trajectory.

\section{Conclusion}
In this paper, we analyze whether ordered visual concepts in T2I diffusion models can be traversed along a single direction, and find that in general they cannot. We therefore describe the traversal as a 1D curve and propose KANSteer, which fits it across all pairwise attribute transitions, keeping a single scalar coordinate while letting the steering vector depend on where the image already sits along the concept. Linear steering is the special case of a straight curve, so the difference between the two measures how far a concept is from a line. That difference is consistent across three FLUX.1 configurations, holds for attributes withheld from fitting, and is largest on the concepts whose trajectories bend the most. Linearity is therefore a local approximation rather than a description of a concept over its full range, and we hope that treating the traversal as a curve offers a useful starting point for interpreting and controlling non-linear concepts in generative models.

\subsection*{AI use statement}
We used a large language model solely to aid in polishing the writing and improving the clarity of the manuscript. The model was not involved in ideation, data analysis, or deriving any of the scientific contributions presented in this work.

\subsection*{Ethics statement}
We acknowledge the potential ethical implications of deploying generative models, including issues related to privacy, data misuse, and the propagation of biases. All models used in this paper are publicly available. We will release the codes to reproduce the results of this paper. We also want to point out the potential role of steering approaches in the generation of fake content, and we encourage and support responsible usage. 

\subsection*{Reproducibility statement}
To facilitate reproducibility, we will make the entire source code and scripts needed to replicate all results presented in this paper available after the peer review period. We conducted all experiments using publicly accessible models. Details of all experiments have been provided in the Appendices. 

\subsubsection*{Acknowledgments}

\bibliography{main}
\bibliographystyle{iclr2027_conference}

\newpage
\appendix
\section{Appendix}

\subsection{Most of the Concepts are not linear in activation space}
\label{app_sec:motivation}

\begin{figure}
    \begin{center}
        \includegraphics[width=0.8\linewidth]{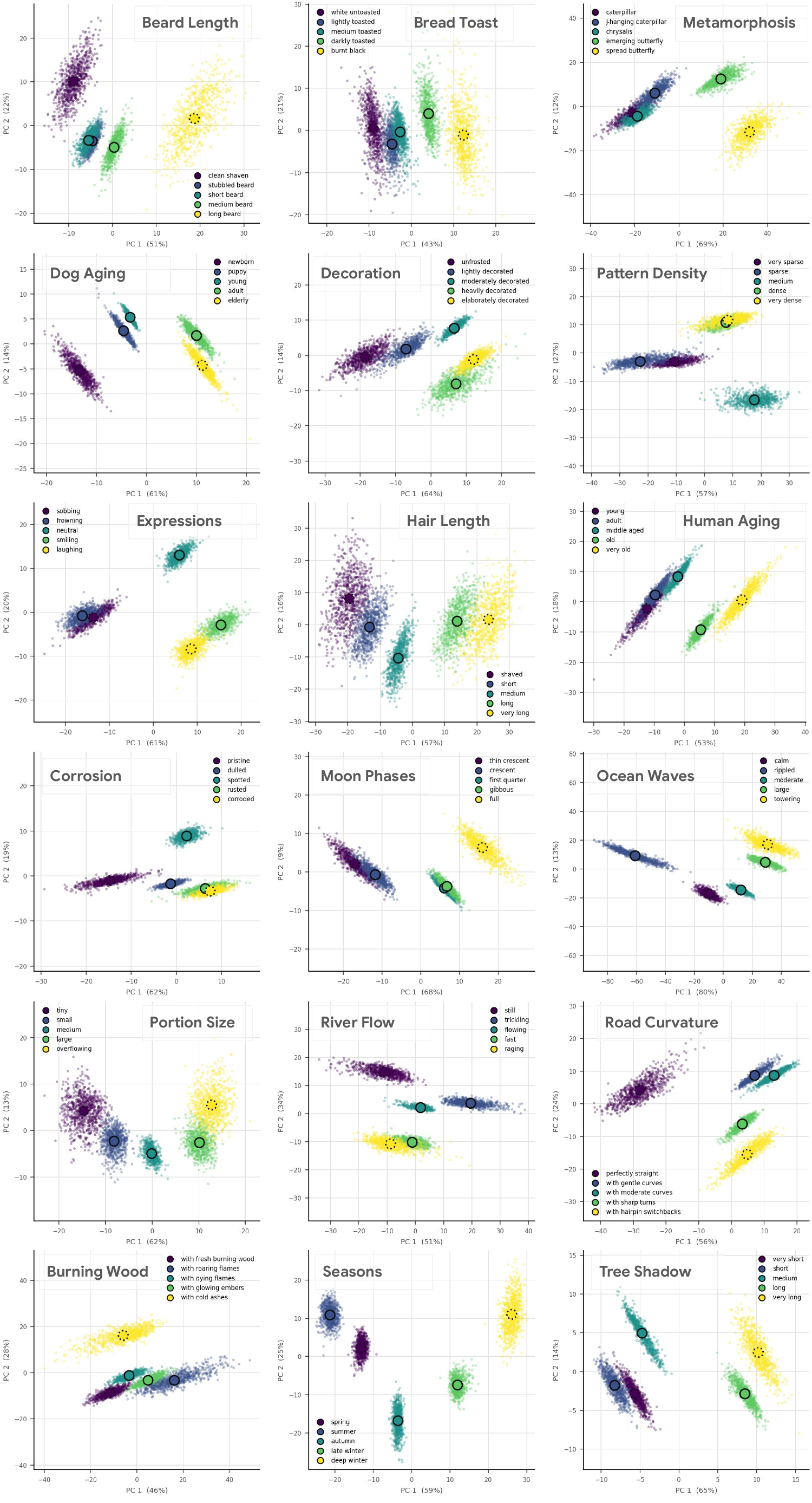}
    \end{center}
    \caption{PCA projections of per-seed, per-attribute delta activations. Each point is one seed; large markers are attribute centroids; the dotted line connects the two endpoints. Intermediate attributes consistently bend away from the endpoint line.}
    \label{fig:pca_app}
\end{figure}

Figure~\ref{fig:motivation} shows PCA projections for three concepts; Figure~\ref{fig:pca_app} extends this to all remaining concepts in our evaluation set. In every case the intermediate attribute centroids deviate from the straight line between the two endpoints, confirming that the nonlinearity is not concept-specific but a general property of how diffusion models encodes ordered attributes. 

\begin{figure}[h]
    \begin{center}
        \includegraphics[width=0.5\linewidth]{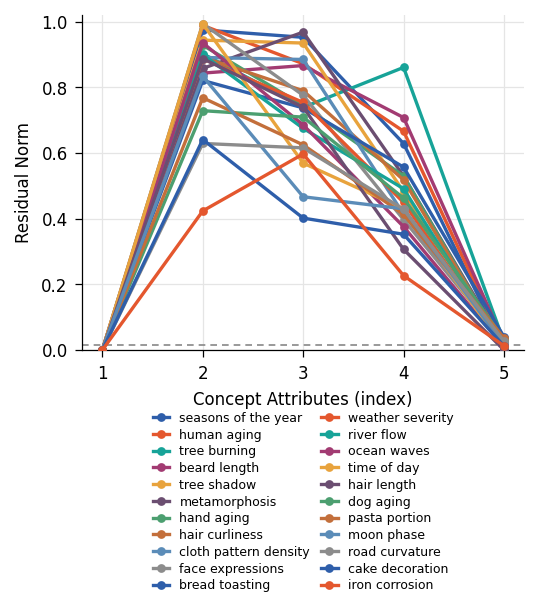}
    \end{center}
    \caption{Residual norm $\rho_k$ per concept. $\rho_k = 0$ means the displacement to attribute $k$ lies exactly on the endpoint direction; $\rho_k = 1$ means it is entirely orthogonal. The dashed line marks the held-out endpoint average. Intermediate attributes (indices 2-4) are far from the line for every concept.}
    \label{fig:chord_dev}
\end{figure}

To quantify how far each intermediate attribute sits from the best possible linear trajectory, we compute the noise-corrected residual norm $\rho_k$ (Eq.~in Sec.~\ref{sec:motivation}). The steering direction $v = \mu_K - \mu_1$ is fitted on a held-out split of 700 seeds, and $\rho_k$ is evaluated on the remaining seeds, so even the endpoint level is a genuine test rather than an identity. Centroid noise is corrected via a half-split estimator where the energy of the noise residual is subtracted from both numerator and denominator so that a perfectly collinear concept returns $\rho_k = 0$ at any sample size.

Figure~\ref{fig:chord_dev} reports $\rho_k$ for all concepts. At every intermediate attribute (indices 2-4) the residual norm exceeds 0.4 and typically exceeds 0.7, meaning that at best 30\% of the displacement to an intermediate level lies along the endpoint direction. This is the best case for linear steering where no choice of scaling $\alpha$ can reduce $\rho_k$ further. The consistently high values across concepts confirm that the nonlinearity observed in the PCA projections is quantitatively large, which ultimately indicates towards the need of a non-linear trajectory fitting method.

\subsection{Localizing Activation Steering Block}
\label{app_sec:localize_block}
We run three interventions on the block-18 update $x_p = H^{19}_p - H^{18}_p$ (Eq.~\ref{eq:delta}). Each sets $x_p$ to zero for a chosen set of patches $S$, so that $H^{19}_p = H^{18}_p$ for $p\in S$, and lets the remaining blocks run unchanged. All generations share the seed and prompt template within a concept.

\begin{figure}[t]
    \begin{center}
        \includegraphics[width=\linewidth]{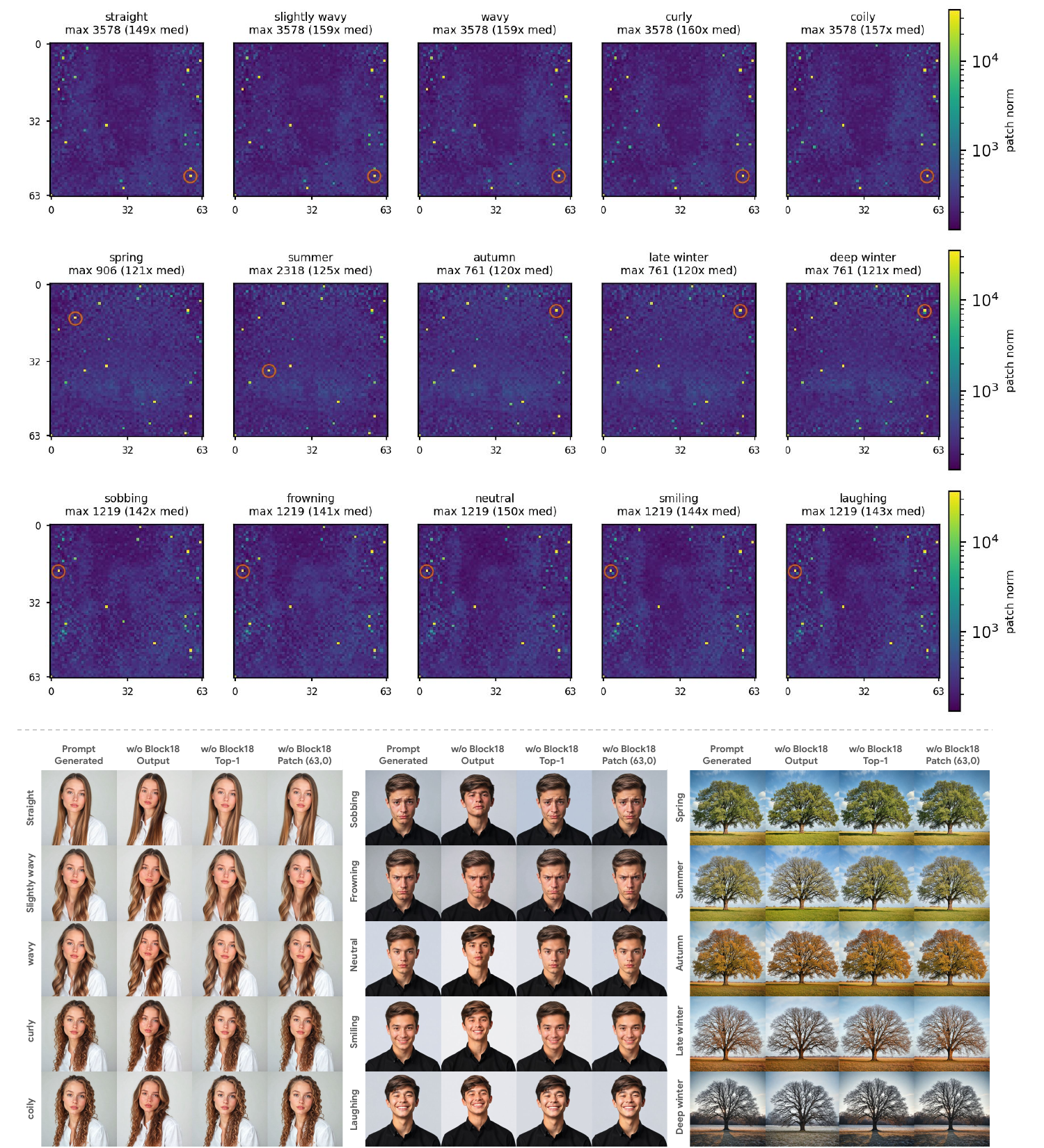}
    \end{center}
    \caption{\textbf{Block 18 of FLUX.1 writes the concept; its massive-norm patches do not.} \textbf{Top:} Patch-norm maps $\lVert x_p\rVert_2$ of the block-18 image-stream delta activations for the five ordered attributes of hair curliness, facial expression and season. A handful of patches sit two orders of magnitude above the median, and for a given concept they occupy nearly the same location across attributes. \textbf{Bottom:} Generations for the same seed and prompt template when (col.~1) nothing is removed, (col.~2) the entire block-18 update is removed, (col.~3) only the top-1\% patches of the update is removed, (col.~4) only the update at patch $(63,0)$ is removed. Removing the block erases the attribute, whereas removing the massive patches leaves the generation, and the attribute, unchanged.}
    \label{fig_app:patch_norm}
\end{figure}

\paragraph{(i) Block 18 adds the concept.} Setting $S$ to all $P=4096$ patches removes the block's entire contribution to the image stream. The resulting images (Fig.~\ref{fig_app:patch_norm}, column \emph{w/o Block-18 Output}) no longer follow the attribute prompt: the attribute-dependent appearance of the prompt-generated images (curl of the hair, expression, season) is not recovered by the later blocks. The residual stream entering block 18 is therefore not sufficient to specify the attribute, and the update that block 18 writes is where the attribute-specific signal is added. This is the causal counterpart of the norm-based localization and is why we fit and steer at this block.

\paragraph{(ii) A few patches carry very large norm, and they are causally inert.} The norm maps in Fig.~\ref{fig_app:patch_norm} (top) are far from uniform. The maximal patch norm is $120$--$160\times$ the median patch norm in every panel, and the extreme patches are isolated single positions on the grid rather than regions. Their locations are governed by the image content and the seed, not by the attribute: for hair curliness and facial expression the top-1 position is identical across all five attributes, and for the season concept it is identical across the last three. One position, $(63,0)$ (flat index $4032$), is special in that it is flagged as an extreme patch in every seed and every concept we inspected, unlike all other positions, which move with the image. Consistent with the register and sink tokens reported in vision transformers~\citep{turri2026few}, these patches do not lie on the object but at positions where the model stores global computation. To test whether they carry anything the generation depends on, we set $S=\{p^\star\}$ with $p^\star=\arg\max_p\lVert x_p\rVert$ (\emph{w/o Top-1}), and $S=\{(63,0)\}$ (\emph{w/o Patch $(63,0)$}). In both cases the generated images are visually indistinguishable from the unablated ones, and in particular the attribute is preserved (Fig.~\ref{fig_app:patch_norm}, columns 3 and 4). Contrast this with (i): removing one of $4096$ patches has no visible effect, whereas removing all of them removes the concept.

\paragraph{(iii) Removing the top $1\%$ is safe, and necessary for fitting.} The objective in Eq.~\ref{eq:rec} is a squared error, so a patch contributes in proportion to $\lVert x_p\rVert^2$, i.e.\ $10^4$ to $2.5\cdot10^4$ times a median patch at the ratios above. As an order-of-magnitude estimate, if a fraction $\epsilon=0.5\%$ of patches has norm $r=140\times$ the median, their share of the total squared norm is $\epsilon r^2/(\epsilon r^2 + 1-\epsilon)\approx 99\%$. Unfiltered, the fit is thus determined by a few dozen patches per image, whose content we show above to be irrelevant to the attribute, and the remainder of the patches, which do carry the concept, contribute little to the loss. We therefore drop, per image, every patch whose norm exceeds the $99$th percentile of the patch norms ($\approx 41$ of $4096$ patches). To check that this discards no concept information, we generate with the update zeroed on exactly this set, $S=\{p:\lVert x_p\rVert>q_{0.99}\}$ (\emph{w/o Top-1\%}). Combined with (ii), this shows that the filtered patches are inert with respect to the concept and that excluding them from the fit removes a numerical artifact of the loss, not signal. In the fit, the percentile is taken on the input norm as the maximum over the $K$ attributes, so a position flagged under any attribute is excluded from all $K$ (Sec.~\ref{sec:experimental_details}).

\subsection{Detailed Quantitative Results}
\label{app_sec:quan_res}

\begin{table*}
\centering
\caption{FLUX.1 Schnell --- Comparison of linear and KAN directions across ten semantic concepts.}
% ============================================================
% ROW 1
% ============================================================

\begin{subtable}[t]{0.49\textwidth}
\centering
\caption{Hair}
\label{tab:hair_1s}
\resizebox{\linewidth}{!}{
\begin{tabular}{llcccc}
\toprule
\textbf{Attribute} & \textbf{Model} &
\textbf{dir-CLIP $\uparrow$} & \textbf{CLIP-I $\uparrow$} &
\textbf{LPIPS $\downarrow$} & \textbf{CLIP-T2I $\uparrow$} \\
\midrule
Straight & Linear & 0 & 0.990 & 0 & 0.4909 \\
& KAN & 0 & 0.999 & 0 & 0.4909 \\
\midrule
Slightly wavy & Linear & 0.083 & 0.889 & 0.252 & 0.4779 \\
& KAN & \textbf{0.192} & \textbf{0.923} & \textbf{0.260} & \textbf{0.5098} \\
\midrule
Wavy & Linear & 0.124 & 0.869 & 0.311 & 0.5081 \\
& KAN & \textbf{0.209} & \textbf{0.911} & \textbf{0.313} & \textbf{0.5352} \\
\midrule
Curly & Linear & 0.187 & 0.880 & 0.360 & 0.5213 \\
& KAN & \textbf{0.271} & \textbf{0.931} & \textbf{0.355} & \textbf{0.5567} \\
\midrule
Coily & Linear & 0.139 & 0.803 & 0.493 & 0.4819 \\
& KAN & \textbf{0.283} & \textbf{0.915} & \textbf{0.386} & \textbf{0.5604} \\
\bottomrule
\end{tabular}}
\end{subtable}
\hfill
\begin{subtable}[t]{0.49\textwidth}
\centering
\caption{Moon Phase}
\label{tab:moon_1s}
\resizebox{\linewidth}{!}{
\begin{tabular}{llcccc}
\toprule
\textbf{Attribute} & \textbf{Model} &
\textbf{dir-CLIP $\uparrow$} & \textbf{CLIP-I $\uparrow$} &
\textbf{LPIPS $\downarrow$} & \textbf{CLIP-T2I $\uparrow$} \\
\midrule
Thin crescent & Linear & 0 & 0.990 & 0 & 0.6584 \\
& KAN & 0 & 0.999 & 0 & 0.6581 \\
\midrule
Crescent & Linear & 0.015 & 0.914 & 0.005 & 0.6541 \\
& KAN & \textbf{0.019} & \textbf{0.973} & \textbf{0} & \textbf{0.659} \\
\midrule
First quarter & Linear & 0.001 & 0.868 & 0.024 & 0.6169 \\
& KAN & \textbf{0.018} & \textbf{0.889} & \textbf{0.009} & \textbf{0.6214} \\
\midrule
Gibbous & Linear & 0.146 & 0.849 & 0.079 & 0.5491 \\
& KAN & \textbf{0.162} & \textbf{0.891} & \textbf{0.031} & \textbf{0.5732} \\
\midrule
Full & Linear & 0.195 & 0.854 & 0.229 & 0.5547 \\
& KAN & \textbf{0.227} & \textbf{0.901} & \textbf{0.041} & \textbf{0.5661} \\
\bottomrule
\end{tabular}}
\end{subtable}

\vspace{0.8em}

% ============================================================
% ROW 2
% ============================================================

\begin{subtable}[t]{0.49\textwidth}
\centering
\caption{Age}
\label{tab:age_1s}
\resizebox{\linewidth}{!}{
\begin{tabular}{llcccc}
\toprule
\textbf{Attribute} & \textbf{Model} &
\textbf{dir-CLIP $\uparrow$} & \textbf{CLIP-I $\uparrow$} &
\textbf{LPIPS $\downarrow$} & \textbf{CLIP-T2I $\uparrow$} \\
\midrule
Preteen & Linear & 0 & 0.960 & 0 & 0.582 \\
& KAN & 0 & 0.996 & 0 & 0.5709 \\
\midrule
Teenage & Linear & 0.014 & 0.816 & 0.121 & 0.5108 \\
& KAN & \textbf{0.070} & \textbf{0.900} & \textbf{0.022} & \textbf{0.5206} \\
\midrule
Young adult & Linear & 0.056 & 0.726 & 0.171 & 0.4151 \\
& KAN & \textbf{0.084} & \textbf{0.852} & \textbf{0.041} & \textbf{0.4283} \\
\midrule
Middle-aged & Linear & 0.194 & 0.619 & 0.223 & 0.4584 \\
& KAN & \textbf{0.270} & \textbf{0.724} & \textbf{0.149} & \textbf{0.4756} \\
\midrule
Elderly & Linear & 0.256 & 0.694 & 0.274 & 0.4992 \\
& KAN & \textbf{0.314} & \textbf{0.784} & \textbf{0.230} & \textbf{0.5446} \\
\bottomrule
\end{tabular}}
\end{subtable}
\hfill
\begin{subtable}[t]{0.49\textwidth}
\centering
\caption{Expression}
\label{tab:expression_1s}
\resizebox{\linewidth}{!}{
\begin{tabular}{llcccc}
\toprule
\textbf{Attribute} & \textbf{Model} &
\textbf{dir-CLIP $\uparrow$} & \textbf{CLIP-I $\uparrow$} &
\textbf{LPIPS $\downarrow$} & \textbf{CLIP-T2I $\uparrow$} \\
\midrule
Sobbing & Linear & 0 & 0.990 & 0 & 0.4743 \\
& KAN & 0 & 0.999 & 0 & 0.4774 \\
\midrule
Frowning & Linear & 0.011 & 0.882 & 0.100 & 0.4885 \\
& KAN & \textbf{0.075} & \textbf{0.884} & \textbf{0.068} & \textbf{0.4952} \\
\midrule
Neutral & Linear & -0.010 & 0.852 & 0.048 & 0.4859 \\
& KAN & \textbf{0.093} & \textbf{0.883} & \textbf{0.007} & \textbf{0.4898} \\
\midrule
Smiling & Linear & 0.160 & 0.884 & 0.188 & 0.4695 \\
& KAN & \textbf{0.184} & \textbf{0.896} & \textbf{0.246} & \textbf{0.4715} \\
\midrule
Laughing & Linear & 0.177 & 0.861 & 0.219 & 0.4341 \\
& KAN & \textbf{0.195} & \textbf{0.881} & \textbf{0.294} & \textbf{0.4569} \\
\bottomrule
\end{tabular}}
\end{subtable}

\vspace{0.8em}

% ============================================================
% ROW 3
% ============================================================

\begin{subtable}[t]{0.49\textwidth}
\centering
\caption{Weather}
\label{tab:weather_1s}
\resizebox{\linewidth}{!}{
\begin{tabular}{llcccc}
\toprule
\textbf{Attribute} & \textbf{Model} &
\textbf{dir-CLIP $\uparrow$} & \textbf{CLIP-I $\uparrow$} &
\textbf{LPIPS $\downarrow$} & \textbf{CLIP-T2I $\uparrow$} \\
\midrule
Bright sunny & Linear & 0 & 0.986 & 0 & 0.662 \\
& KAN & 0 & 1.000 & 0 & 0.6598 \\
\midrule
Pale hazy & Linear & 0.045 & 0.897 & 0.093 & 0.5583 \\
& KAN & \textbf{0.097} & \textbf{0.911} & \textbf{0.062} & \textbf{0.5656} \\
\midrule
Uniform overcast & Linear & 0.209 & 0.867 & 0.476 & 0.6487 \\
& KAN & \textbf{0.281} & \textbf{0.898} & \textbf{0.437} & \textbf{0.6492} \\
\midrule
Dark leaden & Linear & 0.312 & 0.881 & 0.559 & 0.6727 \\
& KAN & \textbf{0.353} & \textbf{0.929} & \textbf{0.527} & \textbf{0.6938} \\
\midrule
Black storm & Linear & 0.249 & 0.724 & 0.742 & 0.6496 \\
& KAN & \textbf{0.374} & \textbf{0.934} & \textbf{0.538} & \textbf{0.7148} \\
\bottomrule
\end{tabular}}
\end{subtable}
\hfill
\begin{subtable}[t]{0.49\textwidth}
\centering
\caption{Season}
\label{tab:season_1s}
\resizebox{\linewidth}{!}{
\begin{tabular}{llcccc}
\toprule
\textbf{Attribute} & \textbf{Model} &
\textbf{dir-CLIP $\uparrow$} & \textbf{CLIP-I $\uparrow$} &
\textbf{LPIPS $\downarrow$} & \textbf{CLIP-T2I $\uparrow$} \\
\midrule
Spring & Linear & 0 & 0.974 & 0 & 0.6401 \\
& KAN & 0 & 0.972 & 0 & 0.6425 \\
\midrule
Summer & Linear & 0.041 & 0.981 & 0.231 & 0.6343 \\
& KAN & \textbf{0.033} & \textbf{0.974} & \textbf{0.256} & \textbf{0.6351} \\
\midrule
Autumn & Linear & 0.155 & 0.969 & 0.193 & 0.5734 \\
& KAN & \textbf{0.098} & \textbf{0.986} & \textbf{0.342} & \textbf{0.5784} \\
\midrule
Late winter & Linear & 0.184 & 0.910 & 0.303 & 0.5612 \\
& KAN & \textbf{0.215} & \textbf{0.965} & \textbf{0.414} & \textbf{0.5690} \\
\midrule
Deep winter & Linear & 0.194 & 0.890 & 0.370 & 0.5338 \\
& KAN & \textbf{0.303} & \textbf{0.934} & \textbf{0.558} & \textbf{0.5878} \\
\bottomrule
\end{tabular}}
\end{subtable}

\vspace{0.8em}

% ============================================================
% ROW 4
% ============================================================

\begin{subtable}[t]{0.49\textwidth}
\centering
\caption{Butterfly}
\label{tab:butterfly_1s}
\resizebox{\linewidth}{!}{
\begin{tabular}{llcccc}
\toprule
\textbf{Attribute} & \textbf{Model} &
\textbf{dir-CLIP $\uparrow$} & \textbf{CLIP-I $\uparrow$} &
\textbf{LPIPS $\downarrow$} & \textbf{CLIP-T2I $\uparrow$} \\
\midrule
Caterpillar & Linear & 0 & 0.958 & 0 & 0.5815 \\
& KAN & 0 & 0.961 & 0 & 0.5805 \\
\midrule
J-hanging caterpillar & Linear & 0.0007 & 0.875 & 0.086 & 0.5702 \\
& KAN & \textbf{0.008} & \textbf{0.886} & \textbf{0.052} & \textbf{0.5930} \\
\midrule
Chrysalis & Linear & 0.031 & 0.832 & 0.054 & 0.5203 \\
& KAN & \textbf{0.054} & \textbf{0.863} & 0.197 & \textbf{0.5340} \\
\midrule
Emerging butterfly & Linear & 0.019 & 0.796 & 0.062 & 0.4769 \\
& KAN & \textbf{0.077} & \textbf{0.817} & 0.206 & \textbf{0.5003} \\
\midrule
Spread butterfly & Linear & 0.012 & 0.788 & 0.068 & 0.4662 \\
& KAN & \textbf{0.108} & \textbf{0.822} & 0.229 & \textbf{0.4821} \\
\bottomrule
\end{tabular}}
\end{subtable}
\hfill
\begin{subtable}[t]{0.49\textwidth}
\centering
\caption{Beard}
\label{tab:beard_1s}
\resizebox{\linewidth}{!}{
\begin{tabular}{llcccc}
\toprule
\textbf{Attribute} & \textbf{Model} &
\textbf{dir-CLIP $\uparrow$} & \textbf{CLIP-I $\uparrow$} &
\textbf{LPIPS $\downarrow$} & \textbf{CLIP-T2I $\uparrow$} \\
\midrule
Clean shaven & Linear & 0 & 0.927 & 0 & 0.4551 \\
& KAN & 0 & 0.907 & 0 & 0.4532 \\
\midrule
Stubbled & Linear & 0.113 & 0.848 & 0.212 & 0.5073 \\
& KAN & \textbf{0.119} & \textbf{0.883} & \textbf{0.195} & \textbf{0.5353} \\
\midrule
Short beard & Linear & 0.092 & 0.846 & 0.276 & 0.5286 \\
& KAN & \textbf{0.185} & \textbf{0.893} & \textbf{0.221} & \textbf{0.5452} \\
\midrule
Medium beard & Linear & 0.129 & 0.864 & 0.350 & 0.5969 \\
& KAN & \textbf{0.222} & \textbf{0.913} & \textbf{0.316} & \textbf{0.6077} \\
\midrule
Long beard & Linear & 0.139 & 0.880 & 0.422 & 0.5781 \\
& KAN & \textbf{0.251} & \textbf{0.937} & \textbf{0.390} & \textbf{0.6367} \\
\bottomrule
\end{tabular}}
\end{subtable}

\vspace{0.8em}

% ============================================================
% ROW 5
% ============================================================

\begin{subtable}[t]{0.49\textwidth}
\centering
\caption{Dog Age}
\label{tab:dogage_1s}
\resizebox{\linewidth}{!}{
\begin{tabular}{llcccc}
\toprule
\textbf{Attribute} & \textbf{Model} &
\textbf{dir-CLIP $\uparrow$} & \textbf{CLIP-I $\uparrow$} &
\textbf{LPIPS $\downarrow$} & \textbf{CLIP-T2I $\uparrow$} \\
\midrule
Newborn & Linear & 0 & 0.945 & 0 & 0.6371 \\
& KAN & 0 & 0.939 & 0 & 0.6338 \\
\midrule
Puppy & Linear & -0.009 & 0.970 & 0.170 & 0.6328 \\
& KAN & \textbf{0.007} & \textbf{0.990} & \textbf{0.090} & \textbf{0.6378} \\
\midrule
Young & Linear & 0.093 & 0.983 & 0.290 & 0.6614 \\
& KAN & \textbf{0.134} & \textbf{0.989} & \textbf{0.170} & \textbf{0.6664} \\
\midrule
Adult & Linear & 0.164 & 0.946 & 0.476 & 0.6121 \\
& KAN & \textbf{0.257} & \textbf{0.973} & \textbf{0.360} & \textbf{0.6362} \\
\midrule
Elderly & Linear & 0.274 & 0.917 & 0.635 & 0.6283 \\
& KAN & \textbf{0.310} & \textbf{0.956} & \textbf{0.435} & \textbf{0.6412} \\
\bottomrule
\end{tabular}}
\end{subtable}
\hfill
\begin{subtable}[t]{0.49\textwidth}
\centering
\caption{Hair Length}
\label{tab:hairlength_1s}
\resizebox{\linewidth}{!}{
\begin{tabular}{llcccc}
\toprule
\textbf{Attribute} & \textbf{Model} &
\textbf{dir-CLIP $\uparrow$} & \textbf{CLIP-I $\uparrow$} &
\textbf{LPIPS $\downarrow$} & \textbf{CLIP-T2I $\uparrow$} \\
\midrule
Shaved & Linear & 0 & 0.955 & 0 & 0.5709 \\
& KAN & 0 & 0.952 & 0 & 0.5760 \\
\midrule
Short & Linear & -0.030 & 0.834 & 0.243 & 0.4614 \\
& KAN & \textbf{0.067} & \textbf{0.875} & \textbf{0.022} & \textbf{0.4714} \\
\midrule
Medium & Linear & 0.046 & 0.862 & 0.086 & 0.4766 \\
& KAN & \textbf{0.200} & \textbf{0.918} & \textbf{0.068} & \textbf{0.5236} \\
\midrule
Long & Linear & 0.038 & 0.837 & 0.110 & 0.4056 \\
& KAN & \textbf{0.176} & \textbf{0.868} & \textbf{0.275} & \textbf{0.4637} \\
\midrule
Very long & Linear & 0.029 & 0.796 & 0.134 & 0.3885 \\
& KAN & \textbf{0.219} & \textbf{0.819} & \textbf{0.289} & \textbf{0.4383} \\
\bottomrule
\end{tabular}}
\end{subtable}
\label{tab:all_attributes_1s}

\end{table*}

% ============================================================================================
% ============================================================================================
% ============================================================================================
% ============================================================================================
% ============================================================================================

\begin{table*}
\centering
\caption{FLUX.1 Schnell 4 steps --- Comparison of linear and KAN directions across ten semantic concepts.}
% ============================================================
% ROW 1
% ============================================================

\begin{subtable}[t]{0.49\textwidth}
\centering
\caption{Hair}
\label{hair_4s}
\resizebox{\linewidth}{!}{
\begin{tabular}{llcccc}
\toprule
\textbf{Attribute} & \textbf{Model} &
\textbf{dir-CLIP $\uparrow$} & \textbf{CLIP-I $\uparrow$} &
\textbf{LPIPS $\downarrow$} & \textbf{CLIP-T2I $\uparrow$} \\
\midrule
Straight & Linear & 0 & 0.939 & 0 & 0.4867 \\
& KAN & 0 & 0.946 & 0 & 0.4742 \\
\midrule
Slightly wavy & Linear & -0.009 & 0.861 & 0.263 & 0.4289 \\
& KAN & \textbf{0.11} & \textbf{0.865} & \textbf{0.168} & \textbf{0.4713} \\
\midrule
Wavy & Linear & 0.025 & 0.833 & 0.318 & 0.4519 \\
& KAN & \textbf{0.13} & \textbf{0.848} & \textbf{0.216} & \textbf{0.4929} \\
\midrule
Curly & Linear & 0.083 & 0.746 & 0.379 & 0.4524 \\
& KAN & \textbf{0.186} & \textbf{0.855} & \textbf{0.36} & \textbf{0.5162} \\
\midrule
Coily & Linear & 0.079 & 0.638 & 0.579 & 0.4181 \\
& KAN & \textbf{0.207} & \textbf{0.673} & \textbf{0.456} & \textbf{0.4463} \\
\bottomrule
\end{tabular}}
\end{subtable}
\hfill
\begin{subtable}[t]{0.49\textwidth}
\centering
\caption{Moon Phase}
\label{tab:moon_4s}
\resizebox{\linewidth}{!}{
\begin{tabular}{llcccc}
\toprule
\textbf{Attribute} & \textbf{Model} &
\textbf{dir-CLIP $\uparrow$} & \textbf{CLIP-I $\uparrow$} &
\textbf{LPIPS $\downarrow$} & \textbf{CLIP-T2I $\uparrow$} \\
\midrule
Thin crescent & Linear & 0 & 0.957 & 0 & 0.6537 \\
& KAN & 0 & 0.937 & 0 & 0.6469 \\
\midrule
Crescent & Linear & 0.019 & 0.943 & 0.008 & 0.6501 \\
& KAN & \textbf{0.021} & \textbf{0.961} & \textbf{0.005} & \textbf{0.6645} \\
\midrule
First quarter & Linear & 0.025 & 0.888 & 0.019 & 0.6209 \\
& KAN & \textbf{0.7} & \textbf{0.901} & \textbf{0.01} & \textbf{0.6426} \\
\midrule
Gibbous & Linear & 0.195 & 0.86 & 0.083 & 0.5745 \\
& KAN & \textbf{0.209} & \textbf{0.899} & \textbf{0.032} & \textbf{0.5747} \\
\midrule
Full & Linear & 0.227 & 0.818 & 0.25 & 0.555 \\
& KAN & \textbf{0.26} & \textbf{0.927} & \textbf{0.041} & \textbf{0.5746} \\
\bottomrule
\end{tabular}}
\end{subtable}

\vspace{0.8em}

% ============================================================
% ROW 2
% ============================================================

\begin{subtable}[t]{0.49\textwidth}
\centering
\caption{Age}
\label{tab:age_4s}
\resizebox{\linewidth}{!}{
\begin{tabular}{llcccc}
\toprule
\textbf{Attribute} & \textbf{Model} &
\textbf{dir-CLIP $\uparrow$} & \textbf{CLIP-I $\uparrow$} &
\textbf{LPIPS $\downarrow$} & \textbf{CLIP-T2I $\uparrow$} \\
\midrule
Preteen & Linear & 0 & 0.96 & 0 & 0.582 \\
& KAN & 0 & 0.944 & 0 & 0.5705 \\
\midrule
Teenage & Linear & 0.035 & 0.856 & 0.075 & 0.4912 \\
& KAN & \textbf{0.1} & \textbf{0.873} & \textbf{0.025} & \textbf{0.5021} \\
\midrule
Young adult & Linear & 0.103 & 0.776 & 0.129 & 0.3673 \\
& KAN & \textbf{0.112} & \textbf{0.802} & \textbf{0.047} & \textbf{0.3792} \\
\midrule
Middle-aged & Linear & 0.137 & 0.69 & 0.109 & 0.4688 \\
& KAN & \textbf{0.253} & \textbf{0.74} & \textbf{0.083} & \textbf{0.4735} \\
\midrule
Elderly & Linear & 0.267 & 0.719 & 0.214 & 0.5693 \\
& KAN & \textbf{0.309} & \textbf{0.736} & \textbf{0.235} & \textbf{0.5788} \\
\bottomrule
\end{tabular}}
\end{subtable}
\hfill
\begin{subtable}[t]{0.49\textwidth}
\centering
\caption{Expression}
\label{tab:expression_4s}
\resizebox{\linewidth}{!}{
\begin{tabular}{llcccc}
\toprule
\textbf{Attribute} & \textbf{Model} &
\textbf{dir-CLIP $\uparrow$} & \textbf{CLIP-I $\uparrow$} &
\textbf{LPIPS $\downarrow$} & \textbf{CLIP-T2I $\uparrow$} \\
\midrule
Sobbing & Linear & 0 & 0.828 & 0 & 0.4778 \\
& KAN & 0 & 0.858 & 0 & 0.4702 \\
\midrule
Frowning & Linear & -0.003 & 0.857 & 0.108 & 0.483 \\
& KAN & \textbf{0.001} & \textbf{0.862} & \textbf{0.072} & \textbf{0.4849} \\
\midrule
Neutral & Linear & -0.029 & 0.816 & 0.053 & 0.5089 \\
& KAN & \textbf{0.017} & \textbf{0.828} & \textbf{0.033} & \textbf{0.5139} \\
\midrule
Smiling & Linear & 0.094 & 0.825 & 0.228 & 0.4784 \\
& KAN & \textbf{0.128} & \textbf{0.845} & \textbf{0.194} & \textbf{0.4921} \\
\midrule
Laughing & Linear & 0.146 & 0.816 & 0.293 & 0.4475 \\
& KAN & \textbf{0.12} & \textbf{0.834} & \textbf{0.224} & \textbf{0.4587} \\
\bottomrule
\end{tabular}}
\end{subtable}

\vspace{0.8em}

% ============================================================
% ROW 3
% ============================================================

\begin{subtable}[t]{0.49\textwidth}
\centering
\caption{Weather}
\label{tab:weather_4s}
\resizebox{\linewidth}{!}{
\begin{tabular}{llcccc}
\toprule
\textbf{Attribute} & \textbf{Model} &
\textbf{dir-CLIP $\uparrow$} & \textbf{CLIP-I $\uparrow$} &
\textbf{LPIPS $\downarrow$} & \textbf{CLIP-T2I $\uparrow$} \\
\midrule
Bright sunny & Linear & 0 & 0.941 & 0 & 0.6735 \\
& KAN & 0 & 0.942 & 0 & 0.6802 \\
\midrule
Pale hazy & Linear & -0.027 & 0.85 & 0.156 & 0.5629 \\
& KAN & \textbf{0.157} & \textbf{0.86} & \textbf{0.12} & \textbf{0.5696} \\
\midrule
Uniform overcast & Linear & 0.12 & 0.833 & 0.556 & 0.6252 \\
& KAN & \textbf{0.201} & \textbf{0.851} & \textbf{0.491} & \textbf{0.6391} \\
\midrule
Dark leaden & Linear & 0.172 & 0.772 & 0.727 & 0.6309 \\
& KAN & \textbf{0.269} & \textbf{0.857} & \textbf{0.638} & \textbf{0.6646} \\
\midrule
Black storm & Linear & 0.206 & 0.692 & 0.958 & 0.6665 \\
& KAN & \textbf{0.262} & \textbf{0.786} & \textbf{0.73} & \textbf{0.7083} \\
\bottomrule
\end{tabular}}
\end{subtable}
\hfill
\begin{subtable}[t]{0.49\textwidth}
\centering
\caption{Season}
\label{tab:season_4s}
\resizebox{\linewidth}{!}{
\begin{tabular}{llcccc}
\toprule
\textbf{Attribute} & \textbf{Model} &
\textbf{dir-CLIP $\uparrow$} & \textbf{CLIP-I $\uparrow$} &
\textbf{LPIPS $\downarrow$} & \textbf{CLIP-T2I $\uparrow$} \\
\midrule
Spring & Linear & 0 & 0.945 & 0 & 0.6221 \\
& KAN & 0 & 0.951 & 0 & 0.6252 \\
\midrule
Summer & Linear & 0.078 & 0.953 & 0.346 & 0.6278 \\
& KAN & \textbf{0.079} & \textbf{0.958} & \textbf{0.306} & \textbf{0.6346} \\
\midrule
Autumn & Linear & 0.092 & 0.946 & 0.239 & 0.5596 \\
& KAN & \textbf{0.122} & \textbf{0.95} & \textbf{0.2} & \textbf{0.5662} \\
\midrule
Late winter & Linear & 0.123 & 0.886 & 0.522 & 0.5346 \\
& KAN & \textbf{0.163} & \textbf{0.945} & \textbf{0.339} & \textbf{0.5423} \\
\midrule
Deep winter & Linear & 0.113 & 0.801 & 0.634 & 0.5504 \\
& KAN & \textbf{0.169} & \textbf{0.82} & \textbf{0.673} & \textbf{0.582} \\
\bottomrule
\end{tabular}}
\end{subtable}

\vspace{0.8em}

% ============================================================
% ROW 4
% ============================================================

\begin{subtable}[t]{0.49\textwidth}
\centering
\caption{Butterfly}
\label{tab:butterfly_4s}
\resizebox{\linewidth}{!}{
\begin{tabular}{llcccc}
\toprule
\textbf{Attribute} & \textbf{Model} &
\textbf{dir-CLIP $\uparrow$} & \textbf{CLIP-I $\uparrow$} &
\textbf{LPIPS $\downarrow$} & \textbf{CLIP-T2I $\uparrow$} \\
\midrule
Caterpillar & Linear & 0 & 0.929 & 0 & 0.5713 \\
& KAN & 0 & 0.935 & 0 & 0.5835 \\
\midrule
J-hanging caterpillar & Linear & -0.009 & 0.873 & 0.039 & 0.5611 \\
& KAN & \textbf{0.005} & \textbf{0.883} & \textbf{0} & \textbf{0.5724} \\
\midrule
Chrysalis & Linear & 0.009 & 0.798 & 0.048 & 0.5067 \\
& KAN & \textbf{0.016} & \textbf{0.826} & \textbf{0.033} & \textbf{0.5149} \\
\midrule
Emerging butterfly & Linear & -0.009 & 0.741 & 0.057 & 0.4722 \\
& KAN & \textbf{0.024} & \textbf{0.789} & \textbf{0.068} & \textbf{0.4865} \\
\midrule
Spread butterfly & Linear & 0.013 & 0.691 & 0.07 & 0.4374 \\
& KAN & \textbf{0.029} & \textbf{0.775} & \textbf{0.097} & \textbf{0.4448} \\
\bottomrule
\end{tabular}}
\end{subtable}
\hfill
\begin{subtable}[t]{0.49\textwidth}
\centering
\caption{Beard}
\label{tab:beard_4s}
\resizebox{\linewidth}{!}{
\begin{tabular}{llcccc}
\toprule
\textbf{Attribute} & \textbf{Model} &
\textbf{dir-CLIP $\uparrow$} & \textbf{CLIP-I $\uparrow$} &
\textbf{LPIPS $\downarrow$} & \textbf{CLIP-T2I $\uparrow$} \\
\midrule
Clean shaven & Linear & 0 & 0.912 & 0 & 0.4479 \\
& KAN & 0 & 0.899 & 0 & 0.4397 \\
\midrule
Stubbled & Linear & 0.009 & 0.814 & 0.194 & 0.4937 \\
& KAN & \textbf{0.114} & \textbf{0.867} & \textbf{0.176} & \textbf{0.5081} \\
\midrule
Short beard & Linear & 0.087 & 0.824 & 0.241 & 0.5140 \\
& KAN & \textbf{0.173} & \textbf{0.873} & \textbf{0.199} & \textbf{0.5318} \\
\midrule
Medium beard & Linear & 0.122 & 0.841 & 0.315 & 0.5867 \\
& KAN & \textbf{0.218} & \textbf{0.908} & \textbf{0.284} & \textbf{0.5941} \\
\midrule
Long beard & Linear & 0.143 & 0.860 & 0.397 & 0.5619 \\
& KAN & \textbf{0.245} & \textbf{0.922} & \textbf{0.354} & \textbf{0.6235} \\
\bottomrule
\end{tabular}}
\end{subtable}

\vspace{0.8em}

% ============================================================
% ROW 5
% ============================================================

\begin{subtable}[t]{0.49\textwidth}
\centering
\caption{Dog Age}
\label{tab:dogage_4s}
\resizebox{\linewidth}{!}{
\begin{tabular}{llcccc}
\toprule
\textbf{Attribute} & \textbf{Model} &
\textbf{dir-CLIP $\uparrow$} & \textbf{CLIP-I $\uparrow$} &
\textbf{LPIPS $\downarrow$} & \textbf{CLIP-T2I $\uparrow$} \\
\midrule
Newborn & Linear & 0 & 0.939 & 0 & 0.6190 \\
& KAN & 0 & 0.944 & 0 & 0.6200 \\
\midrule
Puppy & Linear & -0.003 & 0.964 & 0.210 & 0.6499 \\
& KAN & \textbf{0.004} & \textbf{0.971} & \textbf{0.130} & \textbf{0.6571} \\
\midrule
Young & Linear & 0.086 & 0.967 & 0.350 & 0.6770 \\
& KAN & \textbf{0.149} & \textbf{0.979} & \textbf{0.210} & \textbf{0.6896} \\
\midrule
Adult & Linear & 0.157 & 0.938 & 0.492 & 0.6203 \\
& KAN & \textbf{0.239} & \textbf{0.960} & \textbf{0.395} & \textbf{0.6289} \\
\midrule
Elderly & Linear & 0.284 & 0.921 & 0.655 & 0.6178 \\
& KAN & \textbf{0.326} & \textbf{0.968} & \textbf{0.469} & \textbf{0.6565} \\
\bottomrule
\end{tabular}}
\end{subtable}
\hfill
\begin{subtable}[t]{0.49\textwidth}
\centering
\caption{Hair Length}
\label{tab:hairlength_4s}
\resizebox{\linewidth}{!}{
\begin{tabular}{llcccc}
\toprule
\textbf{Attribute} & \textbf{Model} &
\textbf{dir-CLIP $\uparrow$} & \textbf{CLIP-I $\uparrow$} &
\textbf{LPIPS $\downarrow$} & \textbf{CLIP-T2I $\uparrow$} \\
\midrule
Shaved & Linear & 0 & 0.954 & 0 & 0.5514 \\
& KAN & 0 & 0.959 & 0 & 0.5558 \\
\midrule
Short & Linear & 0.005 & 0.811 & 0.219 & 0.4584 \\
& KAN & \textbf{0.078} & \textbf{0.879} & \textbf{0.196} & \textbf{0.4625} \\
\midrule
Medium & Linear & 0.034 & 0.854 & 0.071 & 0.4649 \\
& KAN & \textbf{0.214} & \textbf{0.901} & \textbf{0.049} & \textbf{0.5074} \\
\midrule
Long & Linear & 0.049 & 0.829 & 0.160 & 0.3849 \\
& KAN & \textbf{0.156} & \textbf{0.844} & \textbf{0.282} & \textbf{0.4512} \\
\midrule
Very long & Linear & 0.063 & 0.762 & 0.156 & 0.3975 \\
& KAN & \textbf{0.228} & \textbf{0.801} & \textbf{0.293} & \textbf{0.4358} \\
\bottomrule
\end{tabular}}
\end{subtable}
\label{tab:all_attributes_4s}

\end{table*}

%%%%%%%%%%%%%%%%%%%%%%%%%%%%%%%%%%%%%%%%%%%%%%%%%%%%%%%%%%%%%%%%%%%
%%%%%%%%%%%%%%%%%%%%%%%%%%%%%%%%%%%%%%%%%%%%%%%%%%%%%%%%%%%%%%%%%%%

\begin{table*}
\centering
\caption{FLUX.1 Dev --- Comparison of linear and KAN directions across ten semantic concepts.}
% ============================================================
% ROW 1
% ============================================================

\begin{subtable}[t]{0.49\textwidth}
\centering
\caption{Hair}
\label{tab:hair_dev}
\resizebox{\linewidth}{!}{
\begin{tabular}{llcccc}
\toprule
\textbf{Attribute} & \textbf{Model} &
\textbf{dir-CLIP $\uparrow$} & \textbf{CLIP-I $\uparrow$} &
\textbf{LPIPS $\downarrow$} & \textbf{CLIP-T2I $\uparrow$} \\
\midrule
Straight & Linear & 0 & 0.857 & 0 & 0.4041 \\
& KAN & 0 & 0.869 & 0 & 0.4327 \\
\midrule
Slightly wavy & Linear & 0.046 & 0.826 & 0.209 & 0.4428 \\
& KAN & \textbf{0.063} & \textbf{0.84} & \textbf{0.16} & \textbf{0.4652} \\
\midrule
Wavy & Linear & 0.06 & 0.814 & 0.256 & 0.4395 \\
& KAN & \textbf{0.097} & \textbf{0.845} & \textbf{0.209} & \textbf{0.4684} \\
\midrule
Curly & Linear & 0.06 & 0.79 & 0.299 & 0.4502 \\
& KAN & \textbf{0.14} & \textbf{0.869} & \textbf{0.255} & \textbf{0.4784} \\
\midrule
Coily & Linear & 0.105 & 0.772 & 0.33 & 0.4443 \\
& KAN & \textbf{0.17} & \textbf{0.798} & \textbf{0.296} & \textbf{0.4923} \\
\bottomrule
\end{tabular}}
\end{subtable}
\hfill
\begin{subtable}[t]{0.49\textwidth}
\centering
\caption{Moon Phase}
\label{tab:moon_dev}
\resizebox{\linewidth}{!}{
\begin{tabular}{llcccc}
\toprule
\textbf{Attribute} & \textbf{Model} &
\textbf{dir-CLIP $\uparrow$} & \textbf{CLIP-I $\uparrow$} &
\textbf{LPIPS $\downarrow$} & \textbf{CLIP-T2I $\uparrow$} \\
\midrule
Thin crescent & Linear & 0 & 0.882 & 0 & 0.6389 \\
& KAN & 0 & 0.888 & 0 & 0.6518 \\
\midrule
Crescent & Linear & -0.014 & 0.85 & 0.019 & 0.6157 \\
& KAN & \textbf{0.006} & \textbf{0.909} & \textbf{0.011} & \textbf{0.6479} \\
\midrule
First quarter & Linear & 0.012 & 0.734 & 0.05 & 0.506 \\
& KAN & \textbf{0.017} & \textbf{0.858} & \textbf{0.021} & \textbf{0.6058} \\
\midrule
Gibbous & Linear & 0.019 & 0.707 & 0.026 & 0.4632 \\
& KAN & \textbf{0.067} & \textbf{0.847} & \textbf{0.199} & \textbf{0.5243} \\
\midrule
Full & Linear & 0.093 & 0.669 & 0.092 & 0.4442 \\
& KAN & \textbf{0.128} & \textbf{0.83} & \textbf{1.071} & \textbf{0.4846} \\
\bottomrule
\end{tabular}}
\end{subtable}

\vspace{0.8em}

% ============================================================
% ROW 2
% ============================================================

\begin{subtable}[t]{0.49\textwidth}
\centering
\caption{Age}
\label{tab:age_dev}
\resizebox{\linewidth}{!}{
\begin{tabular}{llcccc}
\toprule
\textbf{Attribute} & \textbf{Model} &
\textbf{dir-CLIP $\uparrow$} & \textbf{CLIP-I $\uparrow$} &
\textbf{LPIPS $\downarrow$} & \textbf{CLIP-T2I $\uparrow$} \\
\midrule
Preteen & Linear & 0 & 0.849 & 0 & 0.5415 \\
& KAN & 0 & 0.849 & 0 & 0.5415 \\
\midrule
Teenage & Linear & 0.014 & 0.764 & 0.123 & 0.4499 \\
& KAN & \textbf{0.038} & \textbf{0.79} & \textbf{0.088} & \textbf{0.4596} \\
\midrule
Young adult & Linear & 0.037 & 0.71 & 0.14 & 0.3561 \\
& KAN & \textbf{0.084} & \textbf{0.732} & \textbf{0.055} & \textbf{0.3653} \\
\midrule
Middle-aged & Linear & 0.137 & 0.548 & 0.493 & 0.4475 \\
& KAN & \textbf{0.241} & \textbf{0.637} & \textbf{0.142} & \textbf{0.4661} \\
\midrule
Elderly & Linear & 0.259 & 0.593 & 0.634 & 0.5587 \\
& KAN & \textbf{0.306} & \textbf{0.694} & \textbf{0.285} & \textbf{0.5644} \\
\bottomrule
\end{tabular}}
\end{subtable}
\hfill
\begin{subtable}[t]{0.49\textwidth}
\centering
\caption{Expression}
\label{tab:expression_dev}
\resizebox{\linewidth}{!}{
\begin{tabular}{llcccc}
\toprule
\textbf{Attribute} & \textbf{Model} &
\textbf{dir-CLIP $\uparrow$} & \textbf{CLIP-I $\uparrow$} &
\textbf{LPIPS $\downarrow$} & \textbf{CLIP-T2I $\uparrow$} \\
\midrule
Sobbing & Linear & 0 & 0.66 & 0 & 0.4506 \\
& KAN & 0 & 0.66 & \textbf{0} & 0.455 \\
\midrule
Frowning & Linear & 0.005 & 0.696 & 0.246 & 0.452 \\
& KAN & \textbf{0.007} & \textbf{0.753} & \textbf{0.141} & \textbf{0.4685} \\
\midrule
Neutral & Linear & 0.018 & 0.733 & 0.101 & 0.4601 \\
& KAN & \textbf{0.031} & \textbf{0.761} & \textbf{0.037} & \textbf{0.4795} \\
\midrule
Smiling & Linear & 0.115 & 0.807 & 0.175 & 0.474 \\
& KAN & \textbf{0.138} & \textbf{0.813} & \textbf{0.142} & \textbf{0.4874} \\
\midrule
Laughing & Linear & 0.129 & 0.811 & 0.184 & 0.4403 \\
& KAN & \textbf{0.174} & \textbf{0.814} & \textbf{0.159} & \textbf{0.464} \\
\bottomrule
\end{tabular}}
\end{subtable}

\vspace{0.8em}

% ============================================================
% ROW 3
% ============================================================

\begin{subtable}[t]{0.49\textwidth}
\centering
\caption{Weather}
\label{tab:weather_dev}
\resizebox{\linewidth}{!}{
\begin{tabular}{llcccc}
\toprule
\textbf{Attribute} & \textbf{Model} &
\textbf{dir-CLIP $\uparrow$} & \textbf{CLIP-I $\uparrow$} &
\textbf{LPIPS $\downarrow$} & \textbf{CLIP-T2I $\uparrow$} \\
\midrule
Bright sunny & Linear & 0 & 0.899 & 0 & 0.6638 \\
& KAN & 0 & 0.899 & 0 & 0.6694 \\
\midrule
Pale hazy & Linear & 0.014 & 0.828 & 0.049 & 0.5988 \\
& KAN & \textbf{0.104} & \textbf{0.84} & \textbf{0.027} & \textbf{0.6243} \\
\midrule
Uniform overcast & Linear & 0.044 & 0.791 & 0.308 & 0.5292 \\
& KAN & \textbf{0.153} & \textbf{0.826} & \textbf{0.12} & \textbf{0.629} \\
\midrule
Dark leaden & Linear & 0.041 & 0.798 & 0.424 & 0.4556 \\
& KAN & \textbf{0.211} & \textbf{0.842} & \textbf{0.135} & \textbf{0.6109} \\
\midrule
Black storm & Linear & 0.038 & 0.735 & 0.614 & 0.4714 \\
& KAN & \textbf{0.203} & \textbf{0.804} & \textbf{0.355} & \textbf{0.6207} \\
\bottomrule
\end{tabular}}
\end{subtable}
\hfill
\begin{subtable}[t]{0.49\textwidth}
\centering
\caption{Season}
\label{tab:season_dev}
\resizebox{\linewidth}{!}{
\begin{tabular}{llcccc}
\toprule
\textbf{Attribute} & \textbf{Model} &
\textbf{dir-CLIP $\uparrow$} & \textbf{CLIP-I $\uparrow$} &
\textbf{LPIPS $\downarrow$} & \textbf{CLIP-T2I $\uparrow$} \\
\midrule
Spring & Linear & 0 & 0.829 & 0 & 0.5913 \\
& KAN & 0 & 0.824 & 0 & 0.5904 \\
\midrule
Summer & Linear & 0.009 & 0.921 & 0.109 & 0.6142 \\
& KAN & \textbf{0.037} & \textbf{0.927} & \textbf{0.114} & \textbf{0.6289} \\
\midrule
Autumn & Linear & 0.069 & 0.909 & 0.387 & 0.5482 \\
& KAN & \textbf{0.087} & \textbf{0.915} & \textbf{0.15} & \textbf{0.5572} \\
\midrule
Late winter & Linear & 0.133 & 0.89 & 0.372 & 0.5305 \\
& KAN & \textbf{0.207} & \textbf{0.911} & \textbf{0.424} & \textbf{0.5422} \\
\midrule
Deep winter & Linear & 0.12 & 0.73 & 0.577 & 0.5434 \\
& KAN & \textbf{0.223} & \textbf{0.88} & \textbf{0.678} & \textbf{0.5441} \\
\bottomrule
\end{tabular}}
\end{subtable}

\vspace{0.8em}

% ============================================================
% ROW 4
% ============================================================

\begin{subtable}[t]{0.49\textwidth}
\centering
\caption{Butterfly}
\label{tab:butterfly_dev}
\resizebox{\linewidth}{!}{
\begin{tabular}{llcccc}
\toprule
\textbf{Attribute} & \textbf{Model} &
\textbf{dir-CLIP $\uparrow$} & \textbf{CLIP-I $\uparrow$} &
\textbf{LPIPS $\downarrow$} & \textbf{CLIP-T2I $\uparrow$} \\
\midrule
Caterpillar & Linear & 0 & 0.861 & 0 & 0.5811 \\
& KAN & 0 & 0.867 & 0 & 0.5819 \\
\midrule
J-hanging caterpillar & Linear & 0.003 & 0.84 & 0.084 & 0.5654 \\
& KAN & \textbf{0.009} & \textbf{0.855} & \textbf{0.074} & \textbf{0.5676} \\
\midrule
Chrysalis & Linear & 0.012 & 0.764 & 0.127 & 0.4911 \\
& KAN & \textbf{0.017} & \textbf{0.774} & \textbf{0.115} & \textbf{0.5094} \\
\midrule
Emerging butterfly & Linear & 0.061 & 0.753 & 0.149 & 0.4293 \\
& KAN & \textbf{0.147} & \textbf{0.774} & 0.243 & \textbf{0.4831} \\
\midrule
Spread butterfly & Linear & 0.066 & 0.718 & 0.166 & 0.374 \\
& KAN & \textbf{0.198} & \textbf{0.746} & 0.268 & \textbf{0.4309} \\
\bottomrule
\end{tabular}}
\end{subtable}
\hfill
\begin{subtable}[t]{0.49\textwidth}
\centering
\caption{Beard}
\label{tab:beard_dev}
\resizebox{\linewidth}{!}{
\begin{tabular}{llcccc}
\toprule
\textbf{Attribute} & \textbf{Model} &
\textbf{dir-CLIP $\uparrow$} & \textbf{CLIP-I $\uparrow$} &
\textbf{LPIPS $\downarrow$} & \textbf{CLIP-T2I $\uparrow$} \\
\midrule
Clean shaven & Linear & 0 & 0.870 & 0 & 0.4230 \\
& KAN & 0 & 0.869 & 0 & 0.4250 \\
\midrule
Stubbled & Linear & 0.013 & 0.809 & 0.215 & 0.4500 \\
& KAN & \textbf{0.085} & \textbf{0.842} & \textbf{0.195} & \textbf{0.4561} \\
\midrule
Short beard & Linear & 0.056 & 0.741 & 0.239 & 0.4710 \\
& KAN & \textbf{0.139} & \textbf{0.833} & \textbf{0.197} & \textbf{0.4793} \\
\midrule
Medium beard & Linear & 0.097 & 0.795 & 0.298 & 0.4712 \\
& KAN & \textbf{0.188} & \textbf{0.859} & \textbf{0.216} & \textbf{0.4950} \\
\midrule
Long beard & Linear & 0.123 & 0.836 & 0.357 & 0.5188 \\
& KAN & \textbf{0.195} & \textbf{0.891} & \textbf{0.228} & \textbf{0.5329} \\
\bottomrule
\end{tabular}}
\end{subtable}

\vspace{0.8em}

% ============================================================
% ROW 5
% ============================================================

\begin{subtable}[t]{0.49\textwidth}
\centering
\caption{Dog Age}
\label{tab:dogage_dev}
\resizebox{\linewidth}{!}{
\begin{tabular}{llcccc}
\toprule
\textbf{Attribute} & \textbf{Model} &
\textbf{dir-CLIP $\uparrow$} & \textbf{CLIP-I $\uparrow$} &
\textbf{LPIPS $\downarrow$} & \textbf{CLIP-T2I $\uparrow$} \\
\midrule
Newborn & Linear & 0 & 0.922 & 0 & 0.5910 \\
& KAN & 0 & 0.927 & 0 & 0.5890 \\
\midrule
Puppy & Linear & -0.008 & 0.895 & 0.175 & 0.5764 \\
& KAN & \textbf{0.013} & \textbf{0.900} & \textbf{0.113} & \textbf{0.5813} \\
\midrule
Young & Linear & 0.012 & 0.870 & 0.236 & 0.6180 \\
& KAN & \textbf{0.109} & \textbf{0.891} & \textbf{0.179} & \textbf{0.6425} \\
\midrule
Adult & Linear & 0.086 & 0.913 & 0.297 & 0.6187 \\
& KAN & \textbf{0.178} & \textbf{0.955} & \textbf{0.234} & \textbf{0.6610} \\
\midrule
Elderly & Linear & 0.162 & 0.901 & 0.378 & 0.6248 \\
& KAN & \textbf{0.244} & \textbf{0.946} & \textbf{0.245} & \textbf{0.6890} \\
\bottomrule
\end{tabular}}
\end{subtable}
\hfill
\begin{subtable}[t]{0.49\textwidth}
\centering
\caption{Hair Length}
\label{tab:hairlength_dev}
\resizebox{\linewidth}{!}{
\begin{tabular}{llcccc}
\toprule
\textbf{Attribute} & \textbf{Model} &
\textbf{dir-CLIP $\uparrow$} & \textbf{CLIP-I $\uparrow$} &
\textbf{LPIPS $\downarrow$} & \textbf{CLIP-T2I $\uparrow$} \\
\midrule
Shaved & Linear & 0 & 0.942 & 0 & 0.5620 \\
& KAN & 0 & 0.940 & 0 & 0.5600 \\
\midrule
Short & Linear & 0.009 & 0.792 & 0.175 & 0.4312 \\
& KAN & \textbf{0.067} & \textbf{0.816} & \textbf{0.143} & \textbf{0.4358} \\
\midrule
Medium & Linear & 0.027 & 0.784 & 0.196 & 0.4470 \\
& KAN & \textbf{0.160} & \textbf{0.843} & \textbf{0.149} & \textbf{0.4600} \\
\midrule
Long & Linear & 0.081 & 0.790 & 0.254 & 0.4010 \\
& KAN & \textbf{0.187} & \textbf{0.861} & \textbf{0.185} & \textbf{0.4870} \\
\midrule
Very long & Linear & 0.114 & 0.734 & 0.270 & 0.4250 \\
& KAN & \textbf{0.199} & \textbf{0.872} & \textbf{0.203} & \textbf{0.4892} \\
\bottomrule
\end{tabular}}
\end{subtable}
\label{tab:all_attributes_dev}

\end{table*}

%%%%%%%%%%%%%%%%%%%%%%%%%%%%%%%%%%%%%%%%%%%%%%%%%%%%%%%%%%%%%%%%%%%
%%%%%%%%%%%%%%%%%%%%%%%%%%%%%%%%%%%%%%%%%%%%%%%%%%%%%%%%%%%%%%%%%%%

\subsection{Prompts for generating the concepts}
\label{app_sec:prompts}

In this section, we list the concept along with the attributes, and the prompt templates used to compute steering vectors for our experiments. Each concept is specified as a template $T(\cdot)$ into which one attribute is substituted, together with an ordered attribute set $A = (a_1, \dots, a_5)$; the prompt for attribute $k$ is $T(a_k)$, so that consecutive prompts differ only in the substituted attribute. The fixed text in each template holds the subject, pose, framing and background constant across the five attributes.

\paragraph{Facial emotion.}
$T(\cdot)$ = ``a young man with a $\{\cdot\}$ face, wearing black shirt, facing the camera, straight posture, plain background''. \\
$A$ = (\texttt{sobbing}, \texttt{frowning}, \texttt{neutral}, \texttt{smiling}, \texttt{laughing}).

\paragraph{Hair curliness.}
$T(\cdot)$ = ``a young white woman with long $\{\cdot\}$ hair, wearing white shirt, facing the camera, straight posture, plain background''. \\
$A$ = (\texttt{straight}, \texttt{slightly wavy}, \texttt{wavy}, \texttt{curly}, \texttt{coily}).

\paragraph{Tree across seasons.}
$T(\cdot)$ = ``an oak tree in an open field in $\{\cdot\}$, centered, wide shot''. \\
$A$ = (\texttt{spring}, \texttt{summer}, \texttt{autumn}, \texttt{late winter}, \texttt{deep winter}).

\paragraph{Time of day.}
$T(\cdot)$ = ``$\{\cdot\}$ photograph of a straight two-lane asphalt highway cutting through an empty flat plain''. \\
$A$ = (\texttt{bluehour}, \texttt{sunrise}, \texttt{midday}, \texttt{evening}, \texttt{night}).

\paragraph{Moon phase.}
$T(\cdot)$ = ``a telephoto photo of a $\{\cdot\}$ moon, lit from the right side, centered in a black night sky''. \\
$A$ = (\texttt{thin crescent}, \texttt{crescent}, \texttt{first quarter}, \texttt{gibbous}, \texttt{full}).

\paragraph{Human age.}
$T(\cdot)$ = ``$\{\cdot\}$ male, short black hair, wearing a black t-shirt, facing the camera, straight posture, plain grey background''. \\
$A$ = (\texttt{preteen}, \texttt{teenage}, \texttt{young adult}, \texttt{middle-aged}, \texttt{elderly}).

\paragraph{Weather severity.}
$T(\cdot)$ = ``$\{\cdot\}$ skies over a featureless flat grass plain meeting a straight unbroken horizon, no trees, no hills, no buildings, eye-level wide shot, sky filling the upper two thirds of the frame''. \\
$A$ = (\texttt{bright sunny}, \texttt{pale hazy}, \texttt{uniform overcast}, \texttt{dark leaden}, \texttt{black storm}).

\paragraph{Butterfly metamorphosis.}
$T(\cdot)$ = ``a $\{\cdot\}$, on a single horizontal bare twig, centered and filling the frame, macro photo, plain white background, no leaves, no flowers''. \\
$A$ = (\texttt{caterpillar}, \texttt{J-hanging caterpillar}, \texttt{chrysalis}, \texttt{emerging butterfly}, \texttt{spread butterfly}).

\paragraph{Dog aging.}
$T(\cdot)$ = ``a side-view photo of a $\{\cdot\}$ golden retriever standing on a plain grey floor, plain white background, even lighting''. \\
$A$ = (\texttt{newborn}, \texttt{puppy}, \texttt{young}, \texttt{adult}, \texttt{elderly}).

\paragraph{Hair length.}
$T(\cdot)$ = ``a frontal studio portrait of a young man with $\{\cdot\}$ hair, black t-shirt, shoulders square to the camera, plain grey background''. \\
$A$ = (\texttt{shaved}, \texttt{short}, \texttt{medium}, \texttt{long}, \texttt{very-long}).

\paragraph{Beard length.}
$T(\cdot)$ = ``a frontal studio portrait of a young man with a $\{\cdot\}$ beard, black t-shirt, shoulders square to the camera, plain grey background''. \\
$A$ = (\texttt{shaven}, \texttt{stubbled}, \texttt{short}, \texttt{full}, \texttt{bushy}).

\paragraph{Tree shadow.}
$T(\cdot)$ = ``a wide photo of a single oak tree on an open grass field casting a $\{\cdot\}$ shadow, plain sky, fixed viewpoint''. \\
$A$ = (\texttt{faint}, \texttt{soft}, \texttt{clear}, \texttt{long}, \texttt{deep}).

\paragraph{Hand aging.}
$T(\cdot)$ = ``a close-up photo of a single $\{\cdot\}$ human hand held palm-down against a plain grey surface, even lighting''. \\
$A$ = (\texttt{teenage}, \texttt{young}, \texttt{adult}, \texttt{aging}, \texttt{elderly}).

\paragraph{Iron corrosion.}
$T(\cdot)$ = ``a front-view photo of a flat iron plate filling the frame, the metal $\{\cdot\}$, overcast daylight, plain background''. \\
$A$ = (\texttt{pristine}, \texttt{dulled}, \texttt{spotted}, \texttt{rusted}, \texttt{corroded}).

\paragraph{Cake decoration.}
$T(\cdot)$ = ``a front-view photo of a single round cake on a white plate, the cake $\{\cdot\}$, plain grey background, even lighting''. \\
$A$ = (\texttt{plain}, \texttt{lightly-decorated}, \texttt{decorated}, \texttt{heavily-decorated}, \texttt{ornate}).

\paragraph{Tree burning.}
$T(\cdot)$ = ``a wide photo of a single tree on a plain field, the tree $\{\cdot\}$, plain sky, fixed viewpoint''. \\
$A$ = (\texttt{green}, \texttt{wilting}, \texttt{burning}, \texttt{charred}, \texttt{ashen}).

\paragraph{Cloth pattern density.}
$T(\cdot)$ = ``a flat top-view photo of a square fabric swatch filling the frame, printed with a $\{\cdot\}$ pattern, even lighting''. \\
$A$ = (\texttt{sparse}, \texttt{light}, \texttt{moderate}, \texttt{dense}, \texttt{packed}).

\paragraph{Pasta portion.}
$T(\cdot)$ = ``a top-view photo of a white round plate on a plain grey table holding a $\{\cdot\}$ portion of spaghetti, even lighting''. \\
$A$ = (\texttt{tiny}, \texttt{small}, \texttt{medium}, \texttt{large}, \texttt{overflowing}).

\paragraph{River flow.}
$T(\cdot)$ = ``a wide photo of a river running through a plain rocky bank, the water $\{\cdot\}$, overcast daylight, fixed viewpoint''. \\
$A$ = (\texttt{still}, \texttt{trickling}, \texttt{flowing}, \texttt{fast}, \texttt{raging}).

\paragraph{Road curvature.}
$T(\cdot)$ = ``an aerial photo of a single asphalt road crossing an empty green plain, the road $\{\cdot\}$, overcast daylight, fixed viewpoint''. \\
$A$ = (\texttt{straight}, \texttt{gently-curved}, \texttt{curved}, \texttt{winding}, \texttt{hairpin}).

\paragraph{Ocean waves.}
$T(\cdot)$ = ``a wide photo of the open ocean surface under an overcast sky, the water showing $\{\cdot\}$ waves, fixed viewpoint, horizon centered''. \\
$A$ = (\texttt{calm}, \texttt{rippled}, \texttt{moderate}, \texttt{large}, \texttt{towering}).

\paragraph{Bread toasting.}
$T(\cdot)$ = ``a top-view photo of a single slice of plain white bread on a white plate, the bread $\{\cdot\}$, even lighting, plain background''. \\
$A$ = (\texttt{fresh}, \texttt{lightly-toasted}, \texttt{golden}, \texttt{dark-toasted}, \texttt{burnt}).

% \subsection{Higher Degree Polynomials}
% \label{app_sec:higher_degree_poly}

% \begin{figure}[t]
%     \begin{center}
%         \includegraphics[width=\linewidth]{images/app_higher_degree_polynomial.pdf}
%     \end{center}
%     \caption{Higher Degree Polynomials}
%     \label{fig_app:higher_degree}
% \end{figure}

% \begin{figure}[t]
%     \begin{center}
%         \includegraphics[width=\linewidth]{images/app_steering_gain.pdf}
%     \end{center}
%     \caption{Steering Gain}
%     \label{fig_app:steering_gain}
% \end{figure}

\subsection{Qualitative Results}
\label{app_sec:qual_res}
\begin{figure}[h]
    \begin{center}
        \includegraphics[width=0.95\linewidth]{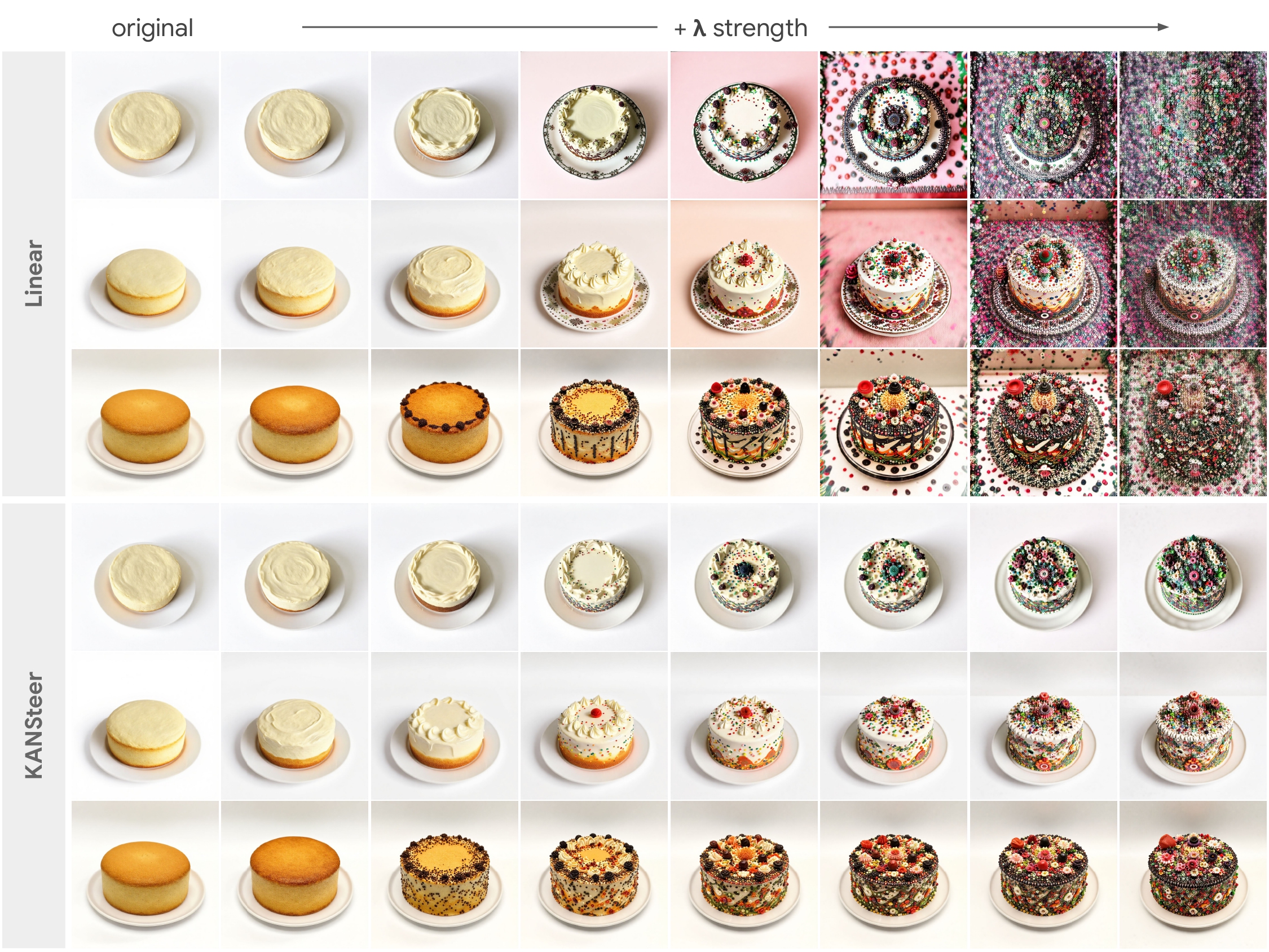}
    \end{center}
    \caption{\textbf{Cake decoration (plain $\rightarrow$ decorated):} Comparison the Linear and KANSteer steering on FLUX.1 Schnell.}
    \label{fig_app:cake_decor}
\end{figure}

\begin{figure}[t]
    \begin{center}
        \includegraphics[width=0.95\linewidth]{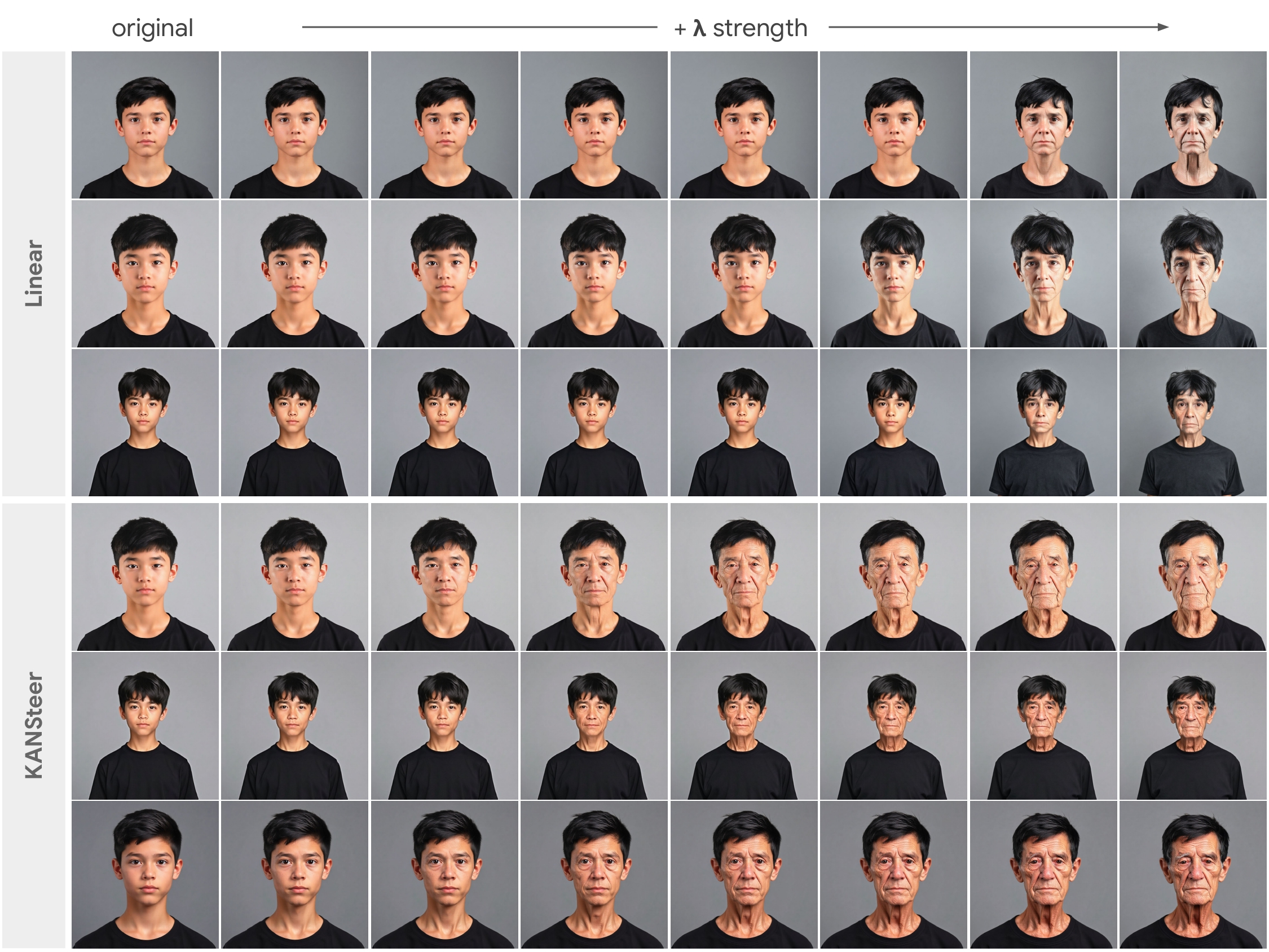}
    \end{center}
    \caption{\textbf{Human Aging (preteen $\rightarrow$ elderly):} Comparison the Linear and KANSteer steering on FLUX.1 Schnell.}
    \label{fig_app:human_aging}
\end{figure}

\begin{figure}[t]
    \begin{center}
        \includegraphics[width=0.95\linewidth]{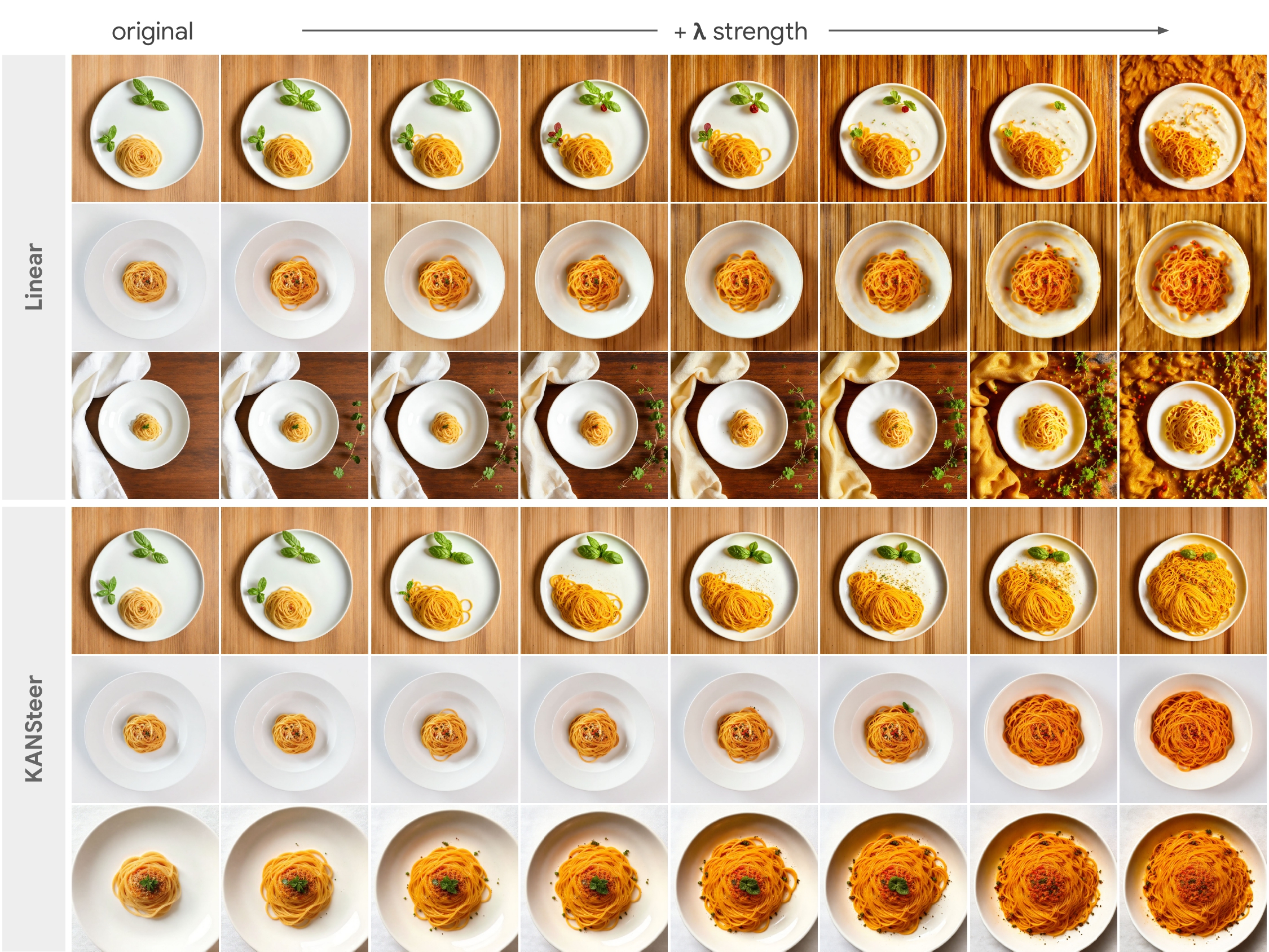}
    \end{center}
    \caption{\textbf{Portion Size (tiny $\rightarrow$ overflowing):} Comparison the Linear and KANSteer steering on FLUX.1 Schnell.}
    \label{fig_app:pasta_portion}
\end{figure}

\begin{figure}[t]
    \begin{center}
        \includegraphics[width=0.95\linewidth]{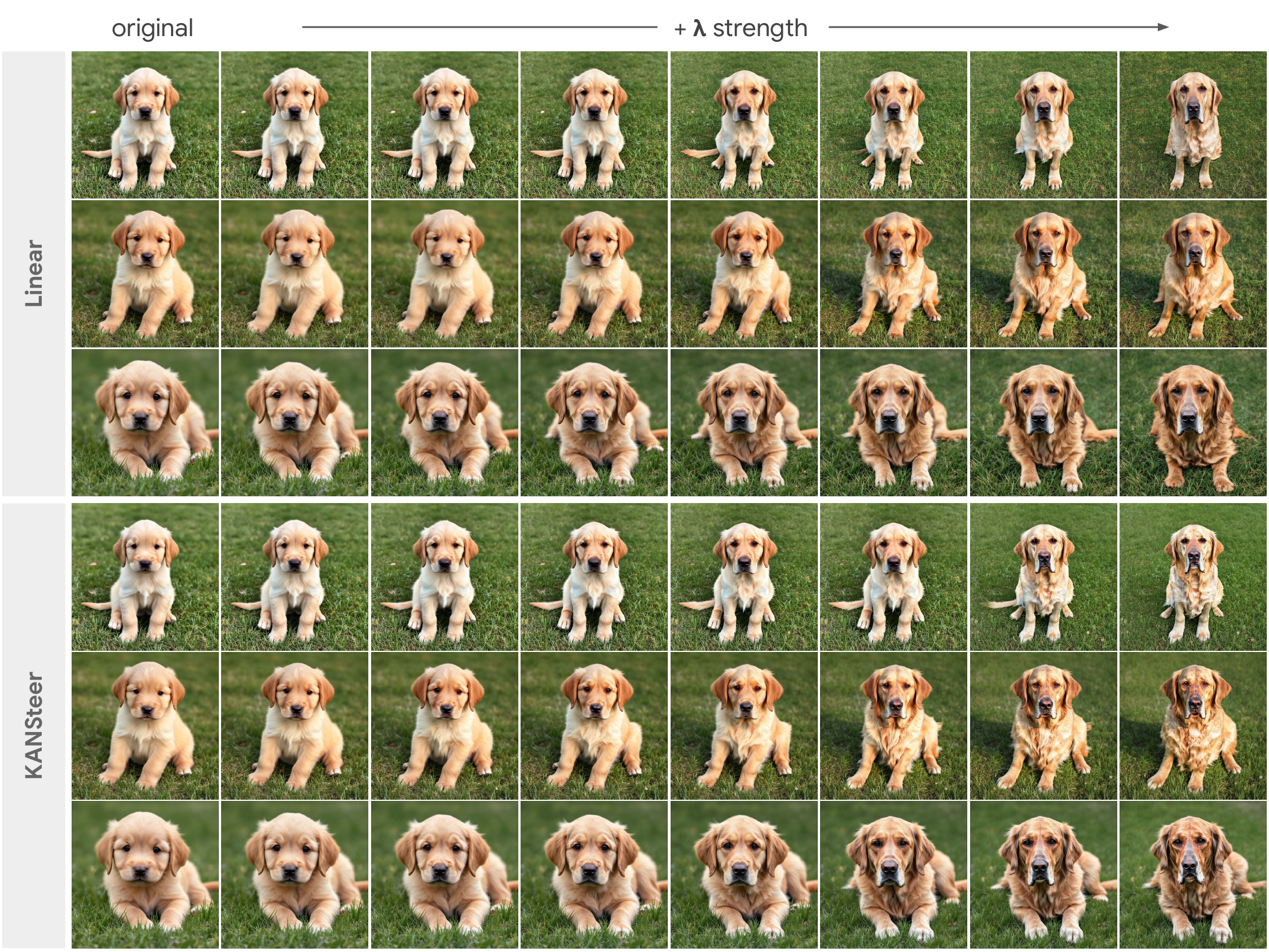}
    \end{center}
    \caption{\textbf{Dog Aging (preteen $\rightarrow$ elderly):} Comparison the Linear and KANSteer steering on FLUX.1 Schnell.}
    \label{fig_app:dog_aging}
\end{figure}

\begin{figure}[t]
    \begin{center}
        \includegraphics[width=0.95\linewidth]{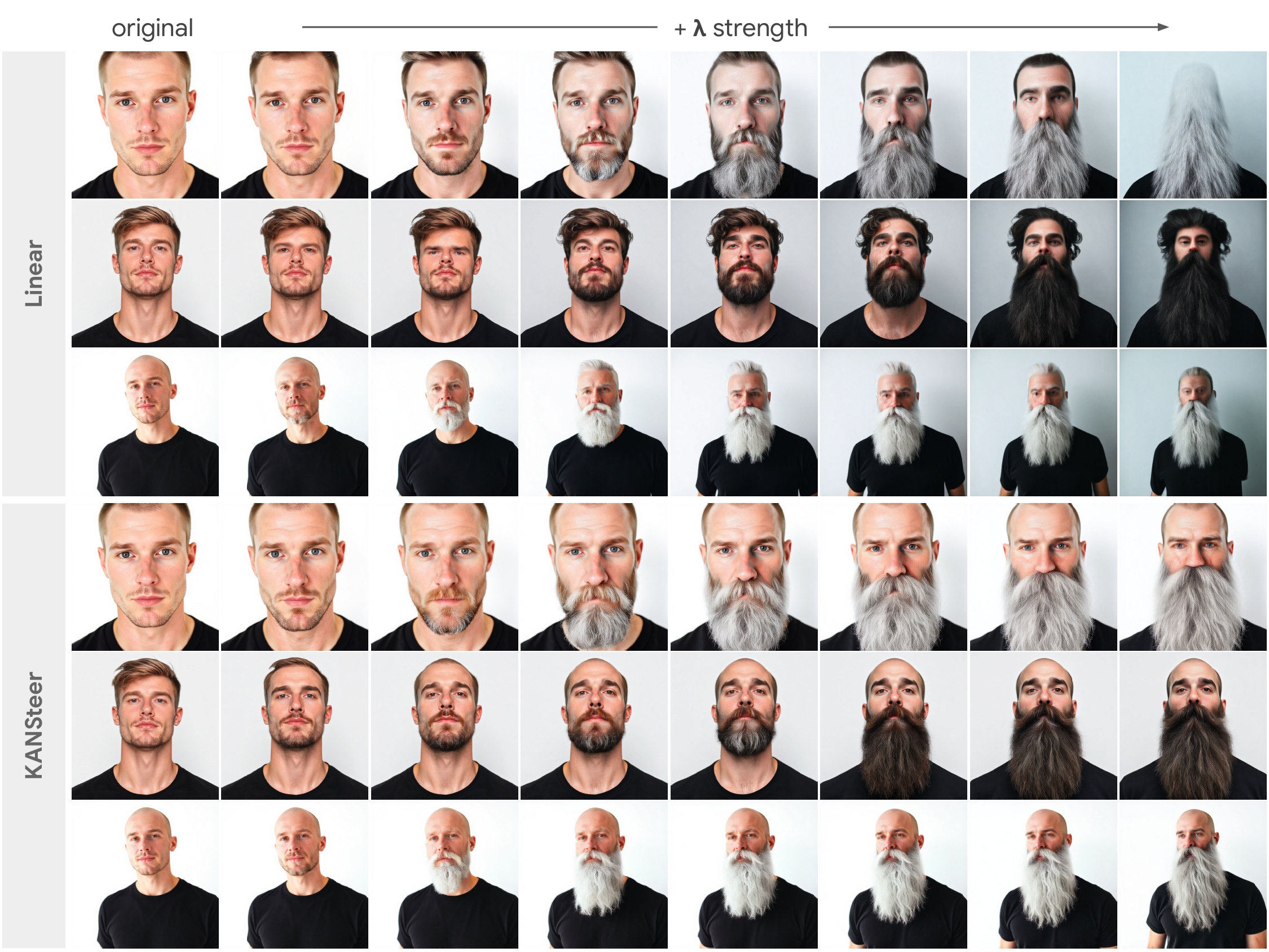}
    \end{center}
    \caption{\textbf{Beard Growth (clean shaven $\rightarrow$ long beard):} Comparison the Linear and KANSteer steering on FLUX.1 Schnell.}
    \label{fig_app:beard}
\end{figure}

\begin{figure}[t]
    \begin{center}
        \includegraphics[width=0.95\linewidth]{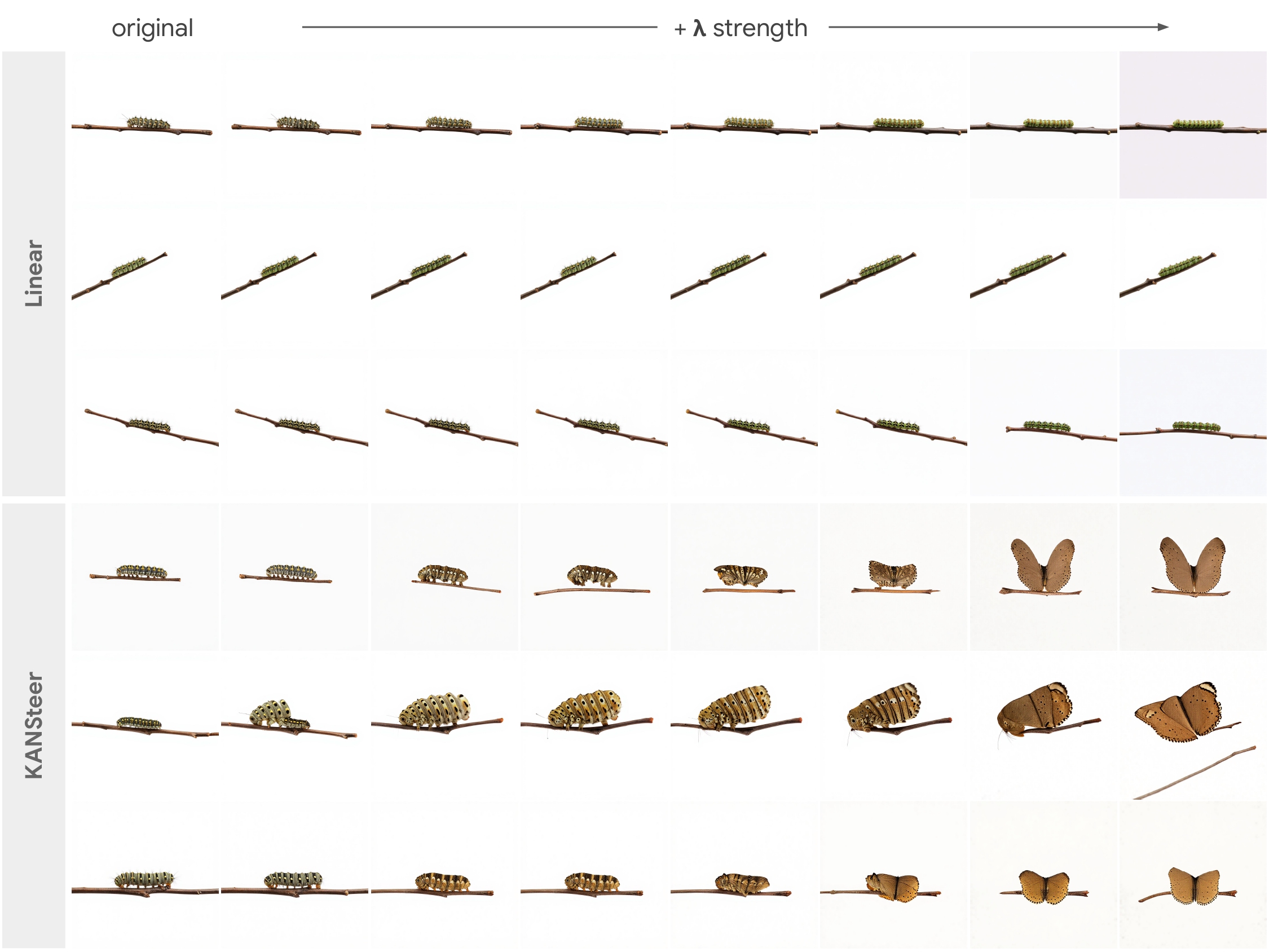}
    \end{center}
    \caption{\textbf{Metamorphosis (caterpillar $\rightarrow$ spread butterfly):} Comparison the Linear and KANSteer steering on FLUX.1 Schnell.}
    \label{fig_app:caterpillar}
\end{figure}

\begin{figure}[t]
    \begin{center}
        \includegraphics[width=0.95\linewidth]{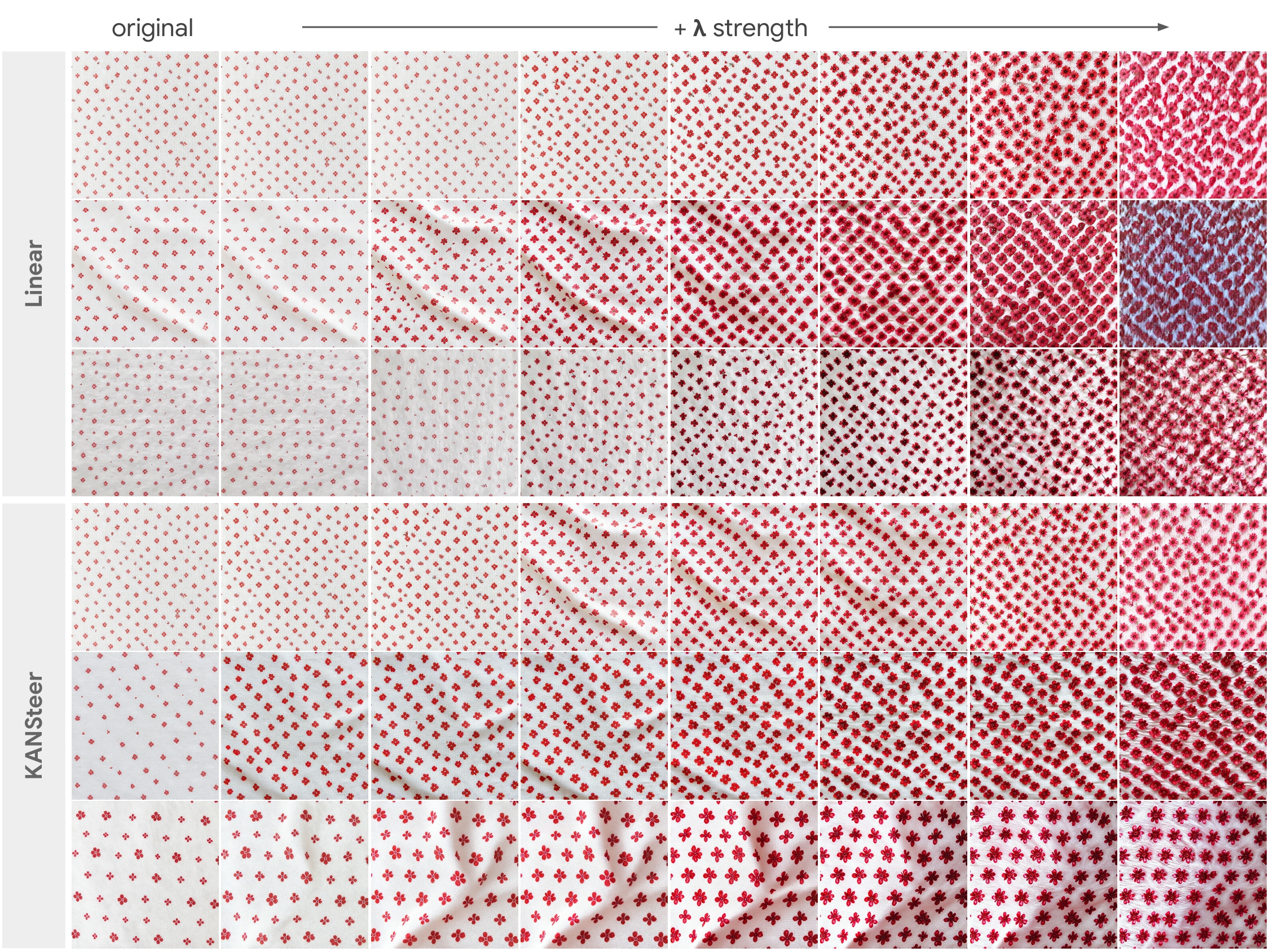}
    \end{center}
    \caption{\textbf{Pattern Density (very sparse $\rightarrow$ very dense):} Comparison the Linear and KANSteer steering on FLUX.1 Schnell.}
    \label{fig_app:pattern_density}
\end{figure}

\begin{figure}[t]
    \begin{center}
        \includegraphics[width=0.95\linewidth]{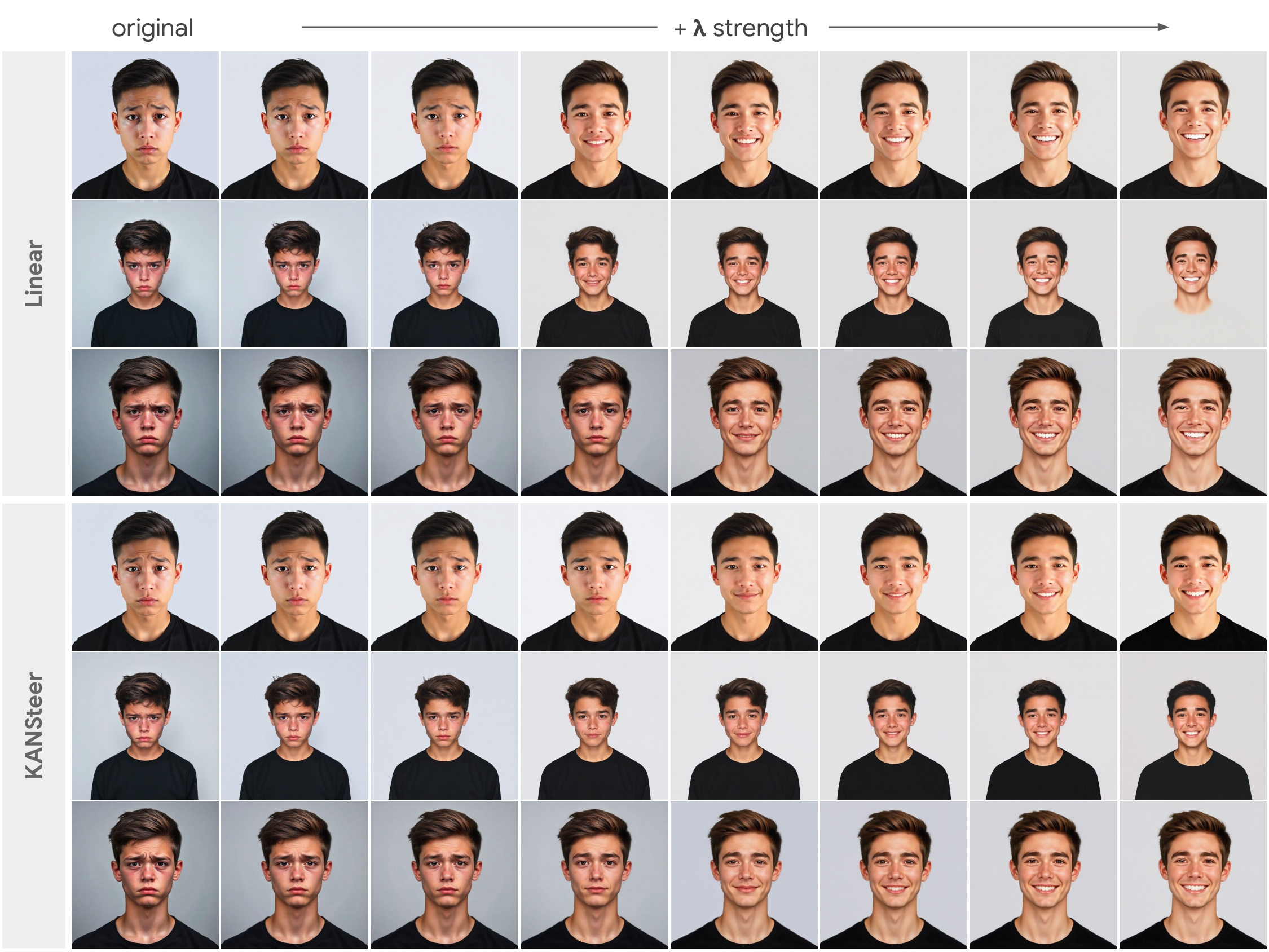}
    \end{center}
    \caption{\textbf{Facial Emotions (sobbing $\rightarrow$ laughing):} Comparison the Linear and KANSteer steering on FLUX.1 Schnell.}
    \label{fig_app:facial_emotion}
\end{figure}

\begin{figure}[t]
    \begin{center}
        \includegraphics[width=0.95\linewidth]{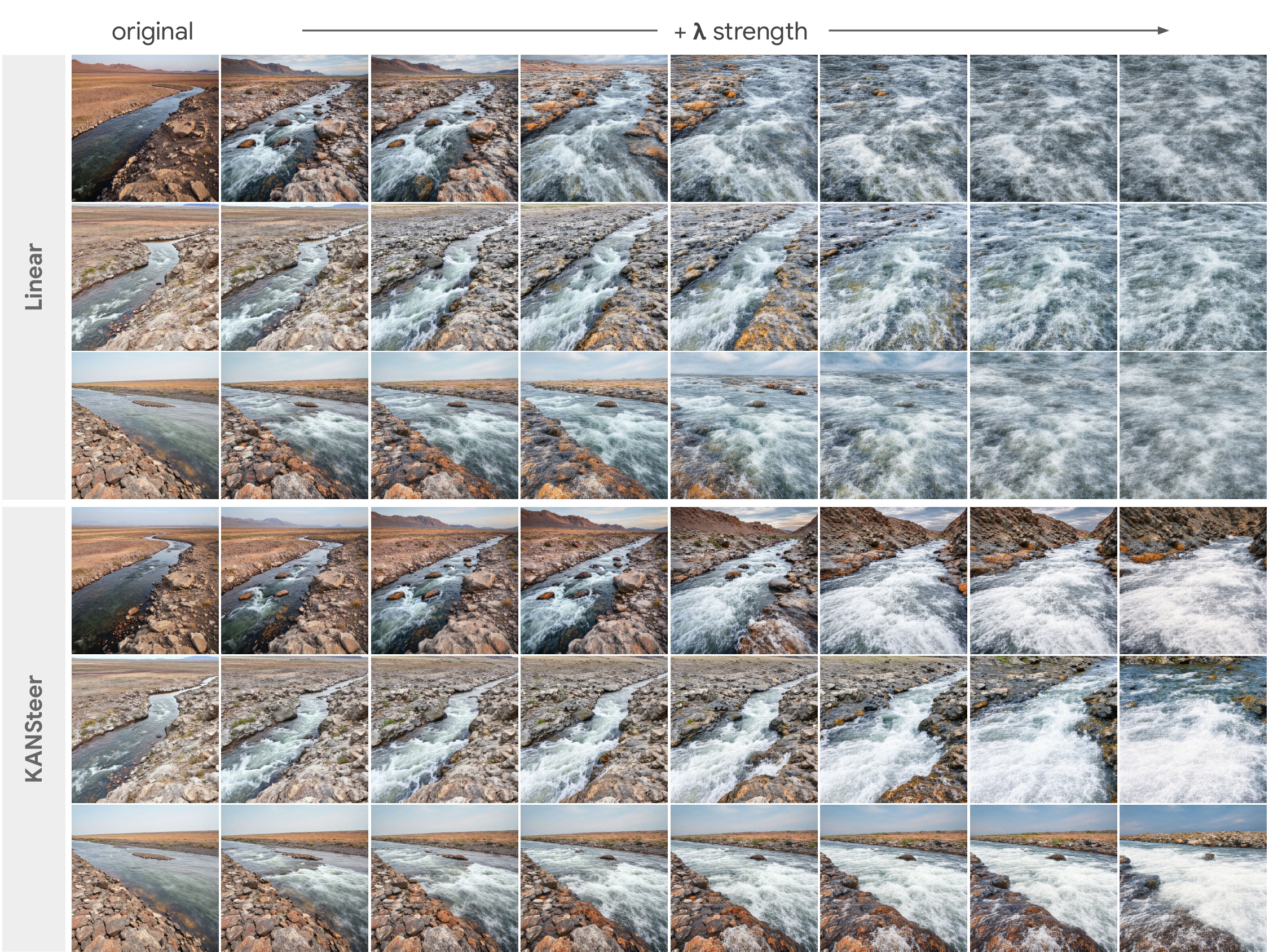}
    \end{center}
    \caption{\textbf{River flow (still $\rightarrow$ raging):} Comparison the Linear and KANSteer steering on FLUX.1 Schnell.}
    \label{fig_app:river}
\end{figure}

\begin{figure}[t]
    \begin{center}
        \includegraphics[width=0.95\linewidth]{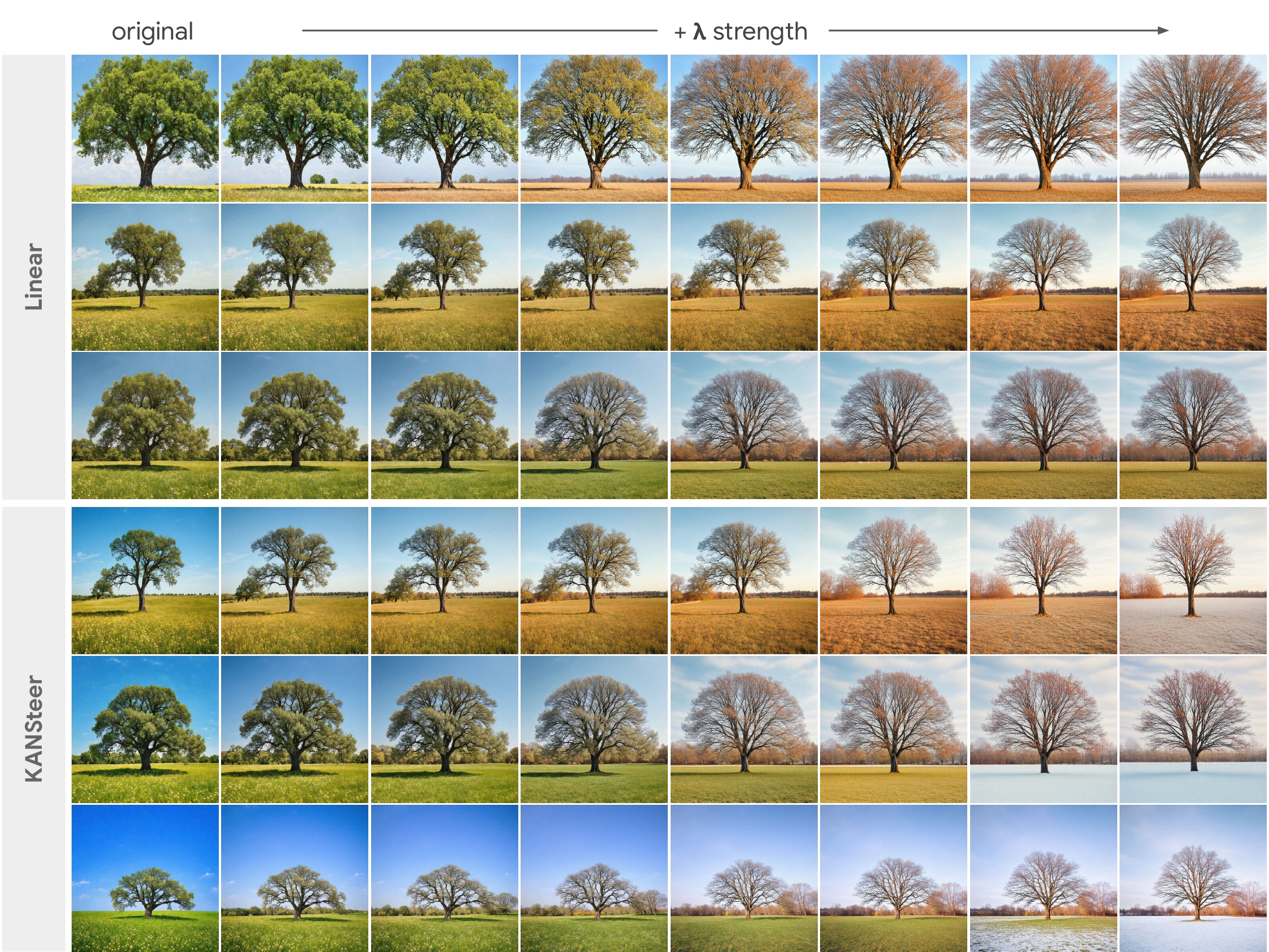}
    \end{center}
    \caption{\textbf{Seasons (spring $\rightarrow$ deep winter):} Comparison the Linear and KANSteer steering on FLUX.1 Schnell.}
    \label{fig_app:season}
\end{figure}

\begin{figure}[t]
    \begin{center}
        \includegraphics[width=0.95\linewidth]{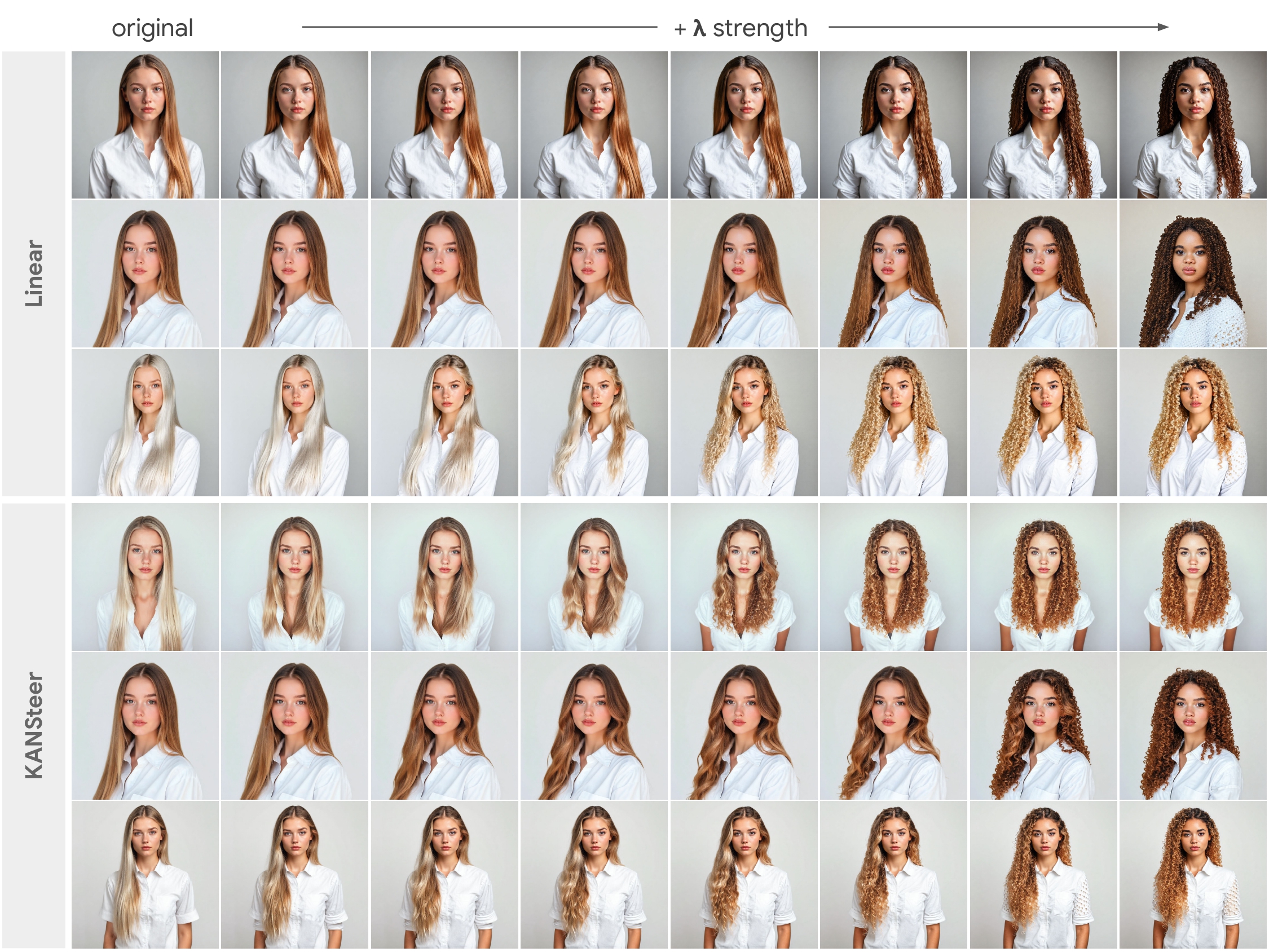}
    \end{center}
    \caption{\textbf{Hair Curliness (straight $\rightarrow$ coily):} Comparison the Linear and KANSteer steering on FLUX.1 Schnell.}
    \label{fig_app:hair_curliness}
\end{figure}

\begin{figure}[t]
    \begin{center}
        \includegraphics[width=0.95\linewidth]{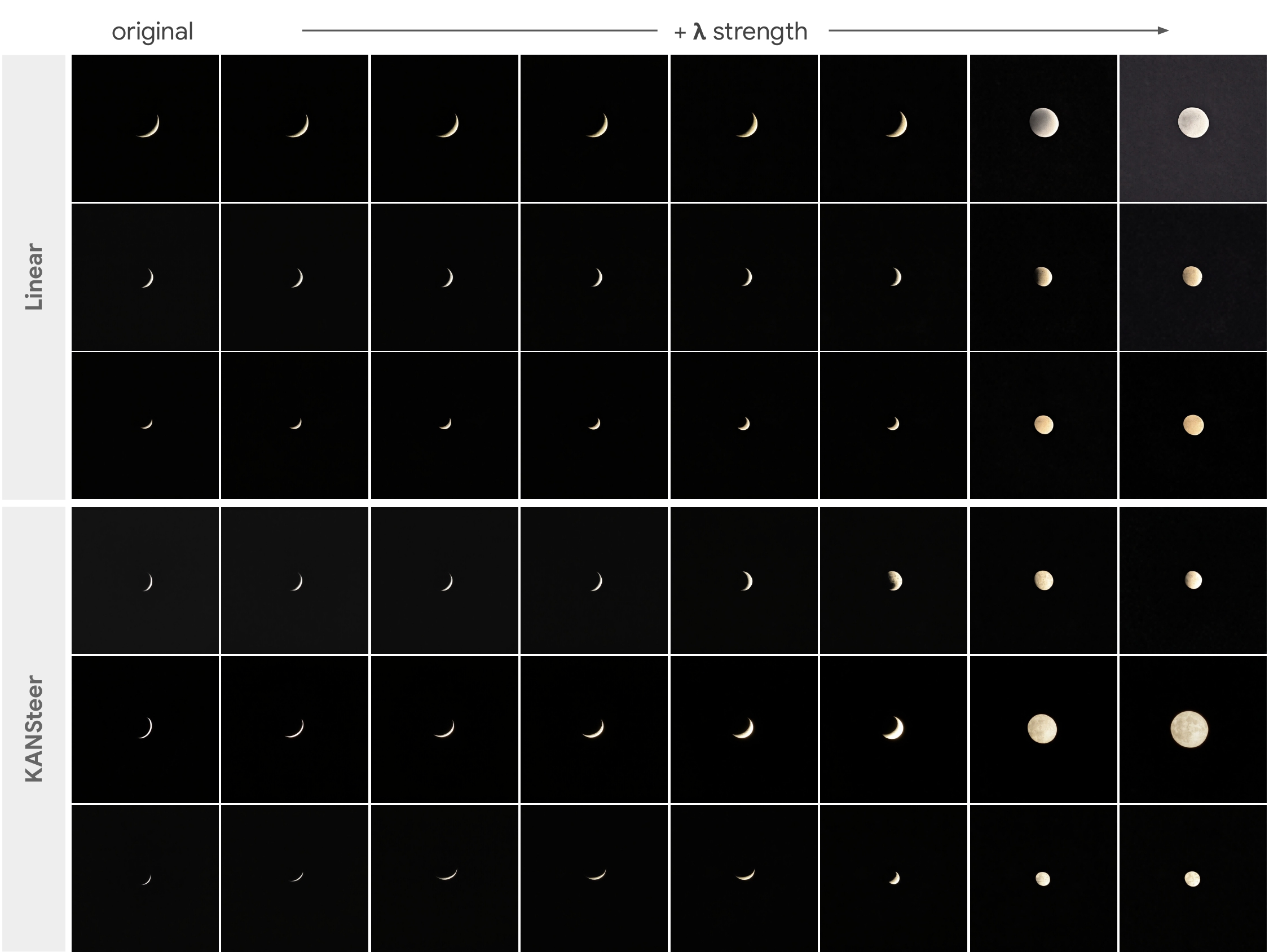}
    \end{center}
    \caption{\textbf{Moon Phases (thin crescent $\rightarrow$ full):} Comparison the Linear and KANSteer steering on FLUX.1 Schnell.}
    \label{fig_app:moon_phases}
\end{figure}

Figures~\ref{fig_app:cake_decor}--\ref{fig_app:moon_phases} show steering sweeps for the concepts on FLUX.1 Schnell. The linear direction and KANSteer are compared under the same conditions, so that within a figure both share the noise, the prompt template and the grid of steering strengths, and the leftmost column is the unsteered generation at $\alpha=0$, which is identical for the two methods by construction. Every curve is fit once on 1-step block-18 activations (Sec.~\ref{sec:experimental_details}) and is applied here without refitting.

\end{document}